\documentclass{article} % For LaTeX2e
\usepackage{iclr2025_conference,times}

\usepackage{amsmath,amsfonts,bm}

\def\eqref#1{equation~\ref{#1}}
\def\1{\bm{1}}

\DeclareMathAlphabet{\mathsfit}{\encodingdefault}{\sfdefault}{m}{sl}
\SetMathAlphabet{\mathsfit}{bold}{\encodingdefault}{\sfdefault}{bx}{n}

\usepackage{xcolor}

\definecolor{lightblue}{RGB}{83,136,233}
\definecolor{blue}{RGB}{32,49,154}

\usepackage[colorlinks, linkcolor=blue, urlcolor=blue,citecolor=lightblue]{hyperref}
\usepackage{url}
\usepackage{booktabs}       % professional-quality tables
\usepackage{amsfonts}       % blackboard math symbols
\usepackage{nicefrac}       % compact symbols for 1/2, etc.
\usepackage{microtype}      % microtypography
\usepackage{wrapfig}
\usepackage{xcolor}         % colors
\usepackage{multirow}
\usepackage{multicol}
\usepackage{adjustbox}
\usepackage{natbib}
\usepackage{subfigure}
\usepackage{graphicx}
\usepackage{subcaption}
\usepackage{pifont}
\newcommand{\tabincell}[2]{\begin{tabular}{@{}#1@{}}#2\end{tabular}}
\newcommand{\cmark}{\ding{51}}%
\newcommand{\xmark}{\ding{55}}%
\usepackage{amsmath}
\usepackage{amsthm}
\title{Quantized Spike-driven Transformer}

\author{Xuerui Qiu$^{1,2,3}$\thanks{Equal Contribution}, \textbf{Malu Zhang}$^{1}$\thanks{Corresponding author, maluzhang@uestc.edu.cn}, Jieyuan Zhang$^{1*}$, Wenjie Wei$^{1}$, Honglin Cao$^{1}$,  \\ \textbf{Junsheng Guo$^{4}$}, \textbf{Rui-Jie Zhu}$^{5}$, \textbf{Yimeng Shan}$^{6}$, \textbf{Yang Yang}$^{1}$,  \textbf{Haizhou Li}$^{7}$ \\ 
~\\
$^{1}$University of Electronic Science and Technology of China,\\
$^{2}$Institute of Automation, Chinese Academy of Sciences,\\
$^{3}$School of Future Technology, University of Chinese Academy of Sciences,\\
$^{4}$China Agricultural University,
$^{5}$University of California, Santa Cruz,\\
$^{6}$Liaoning Technical University,
$^{7}$Chinese University of Hong Kong (Shenzhen)\\
}

\iclrfinalcopy

\begin{document}

\maketitle

\begin{abstract}
Spiking neural networks (SNNs) are emerging as a promising energy-efficient alternative to traditional artificial neural networks (ANNs) due to their spike-driven paradigm.
However, recent research in the SNN domain has mainly focused on enhancing accuracy by designing large-scale Transformer structures, which typically rely on substantial computational resources, limiting their deployment on resource-constrained devices.
To overcome this challenge, we propose a quantized spike-driven Transformer baseline (QSD-Transformer), which achieves reduced resource demands by utilizing a low bit-width parameter. 
Regrettably, the QSD-Transformer often suffers from severe performance degradation.
In this paper, we first conduct empirical analysis and find that the bimodal distribution of quantized spike-driven self-attention (Q-SDSA) leads to spike information distortion (SID) during quantization, causing significant performance degradation. To mitigate this issue, we take inspiration from mutual information entropy and propose a bi-level optimization strategy to rectify the information distribution in Q-SDSA.
Specifically, at the lower level, we introduce an information-enhanced LIF to rectify the information distribution in Q-SDSA.
At the upper level, we propose a fine-grained distillation scheme for the QSD-Transformer to align the distribution in Q-SDSA with that in the counterpart ANN.
By integrating the bi-level optimization strategy, the QSD-Transformer can attain enhanced energy efficiency without sacrificing its high-performance advantage.
We validate the QSD-Transformer on various visual tasks, and experimental results indicate that our method achieves state-of-the-art results in the SNN domain.
For instance, when compared to the prior SNN benchmark on ImageNet, the QSD-Transformer achieves 80.3\% top-1 accuracy, accompanied by significant reductions of 6.0$\times$ and 8.1$\times$ in power consumption and model size, respectively. Code is available at \href{https://github.com/bollossom/QSD-Transformer}{Quantized Spike-driven Transformer.}
\end{abstract}

\section{Introduction}
% \begin{wrapfigure}[13]{r}{0.25\textwidth}
% \begin{center}
%     \includegraphics[width=0.24\textwidth]{bubble.pdf}
%       \end{center}
%     %{-3mm}
%   \caption{Accuracy vs. Power vs. Params. Our QSD-Transformer enjoys lower power and storage savings while surpassing the best results.}
%   %{-3mm}
%   \label{fig:speed-size-acc}
% \end{wrapfigure}
Spiking neural networks (SNNs) have emerged as a promising approach for realizing energy-efficient computational intelligence due to their high biological plausibility \citep{Maass_1997_LIF}, sparse spike-driven communication \citep{Nature_2}, and low power consumption on neuromorphic hardware \citep{loihi,Tianjic,TrueNorth}.
Within SNNs, the spiking neuron transmits information via sparse binary spikes, where the binary value of 0 denotes neural quiescence and the value of 1 signifies a spiking event \citep{Slayer,eshraghian2023training}.  
The unique spike-driven nature is key to achieving low power consumption, where only a subset of spiking neurons are activated to perform sparse synaptic accumulation (AC) \citep{yao2024spike,yao2023spike}. However, despite their high energy efficiency, the application of SNNs is constrained by their low task accuracy.
\par
Numerous researchers have made great efforts to improve the performance and expand the application scenarios of SNNs. Building upon the success of Vision Transformers (ViT) \citep{dosovitskiy2020image,touvron2021training,yu2023metaformer}, researchers naturally combined SNNs with Transformers, resulting in significant performance improvements on ImageNet benchmark \citep{zhou2023spikformer,zhou2023spikingformer,zhou2024qkformer,zhou2024direct} and diverse application scenarios \citep{zhang2022spiking2,zhang2022spike,lv2023spikebert}. Despite their commendable performance, these studies come at the cost of massive model
parameters and high computational complexity. For instance, Spikformer v2 \citep{zhou2024spikformer} and Spike-driven Transformer v3 \cite{yao2024scaling} achieve accuracies of 82.4\% and 86.2\% on the ImageNet dataset, respectively. These models have 173M parameters, necessitating 1384MB memory, and requiring 28.4G synapse-operations per second for inference.
% And Spike-driven Transformer v3 \cite{} achieves an accuracy of 86.2\% on the ImageNet dataset, having 173M parameters, 
This places significant demands on the storage and computational capabilities of neuromorphic chips, thereby limiting their deployment on edge devices. Therefore, there is an urgent need for a low-bit and high-performance Spike-based Transformer.
% These demands significantly strain the memory and computational capabilities of neuromorphic chips. Hence, there arises an urgent need for a high-performance and low-bit Spikformer.
 % As an example, the latest Loihi2 chip \citep{shrestha2024efficient} can only support models with up to approximately 15 M parameters in memory. 
 \par
Numerous efforts have been made to compress and accelerate neural networks on edge computing devices, e.g., pruning \citep{han2015learning,shen2023esl}, quantization \citep{qin2021bibert, deng2021compress}, and knowledge distillation \citep{hinton2015distilling, xu2023constructing}. Among these, quantization is particularly suitable for hardware deployment as it can reduce the bit-width of network parameters and activations, enabling efficient inference. The post-training quantization (PTQ) approach \citep{jacob2018quantization} calculates quantization parameters directly based on pre-trained full-precision models, which may limit the model's performance to a suboptimal level without fine-tuning. In particular, the model obtained from this approach may suffer from dramatic performance drops when quantized to ultra-low bits (e.g., 2, 4 bit). In contrast to PTQ, Quantization-Aware Training (QAT) \citep{krishnamoorthi2018quantizing} performs quantization during the learning process and generally achieves great performance with high compression ratios. 
However, in the field of SNNs, research on QAT methods has primarily focused on convolutional neural networks (CNNs), with low-bit Spikformer remaining largely unexplored.
% Although QAT methods have demonstrated effectiveness in compressing Conv-based SNNs for computer vision tasks, there has been no exploration of low-bit Spikformer.
\par
% In this paper, we first construct a low-bit spike-driven Transformer (SD-Transformer) baseline \citep{yao2024spike}, which is a simple yet efficient solution based on QAT methods. Our investigation of the baseline model revealed a notable performance decline on the ImageNet benchmark. As shown in Fig. \ref{fig:degrad}, applying LSQ \citep{bhalgat2020lsq+} to quantize the weights of a 15M SD-Transformer into 4 bits resulted in only 67.6\% accuracy, exhibiting a 4.2\% gap compared to full precision performance. In contrast, quantizing the same architecture  ANN showed a mere 1.5\% performance gap. We attribute the significant performance drop primarily to spike information distortion in the forward process of the self-attention mechanism, as well as optimization in eliminating gradient distribution discrepancies during backpropagation.
\par
\begin{figure}[!t]
    \centering
    \includegraphics[width=0.95\linewidth]{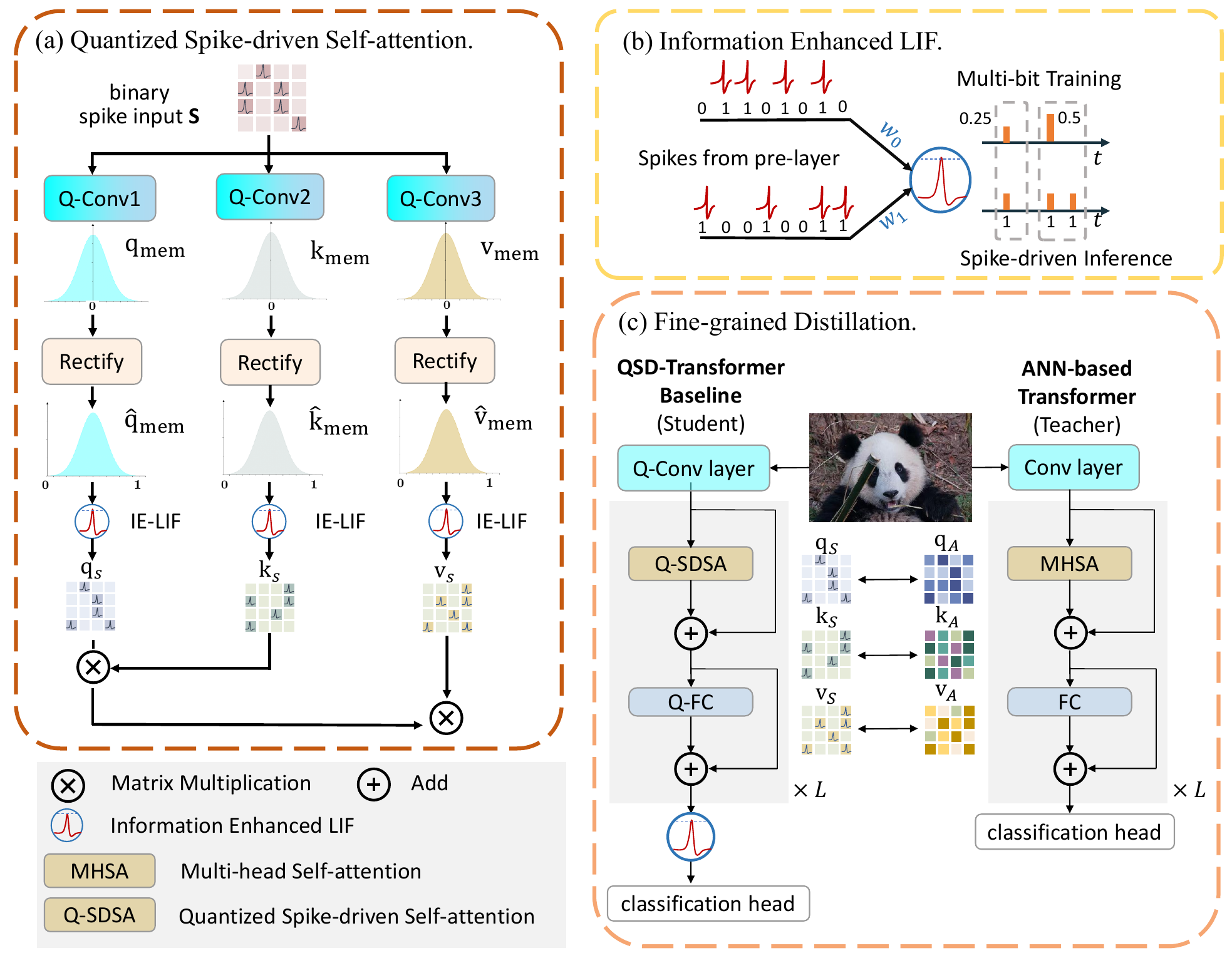}
    \caption{Overview of the QSD-Transformer. (a) Proposed quantized spike-driven self-attention (Q-SDSA) module, where the membrane potential is rectified and then sent to the information-enhanced LIF (IE-LIF) neuron. (b) Proposed IE-LIF spiking neuron model, which utilizes the multi-bit spike during training while the binary spike during inference. (c) Proposed fine-grained distillation scheme.}
    \label{fig:method}
\end{figure}
%我们开发了 a fully quantized spike-driven Transformer (QSD-Transformer)，将全精度纯加法驱动的SD-Transformer转化为低比特模型。通过保持脉冲驱动注意力模块的分布与同架构ANN全精度对应物的分布相同（见图2的概述）。
%在前向过程中我们提出了information-enhanced Spiking Neuron 弥补量化SDSA模块上的脉冲信息失真, which activates Multibit values during training while maintaining spike-driven by extending virtual timestep during inference.在反向传播中，我们提出了一种fine-grained distillation（DGD）方案，通过对QSD-Transofmer和同架构ANN注意力分布之间的差异进行优化，in eliminating gradient distribution discrepancies during backpropagation.
%因此，我们提出通过基于信息熵最大化的信息校正模块（IRM）修改量化注意力模块上的失真分布，在前向过程中。而在反向过程中，我们提出了一种分布引导蒸馏（DGD）方案，通过量化ViT和全精度对应物之间的注意力相似性损失来消除分布变化。我们工作的贡献包括：

% To tackle the above issue, we developed a fully quantized spike-driven Transformer (QSD-Transformer). In our study, we introduce information-enhanced spike neurons (IE-LIF) during the forward process, and in the subsequent backpropagation stage, we propose a fine-grained distillation (FGD) scheme. By integrating these two approaches, we aim to address the challenge of the spike information distortion problem in the quantized spike-driven self-attention  (Q-SDSA) module. Our main contributions can be summarized as:

In this paper, we first construct a quantized spike-driven Transformer (QSD-Transformer) baseline \citep{yao2024spike}, which directly quantizes 32-bit weights to low bit-width during training. 
Despite exhibiting significant energy efficiency, this simple method can lead to severe performance degradation.
Through detailed analysis of the baseline, we reveal that quantizing the attention module will reduce the representation capability of the self-attention maps, which is defined as the spiking information distortion (SID) problem. This is the main reason for the performance degradation.
To address this issue, we propose a bi-level optimization strategy for the baseline, aiming at rectifying the distribution in quantized spike-driven self-attention maps (Q-SDSA) from both the neuron and network levels.
The overview of the QSD-Transformer is shown in Fig. \ref{fig:method} and our main contributions can be summarized as:
\begin{itemize}
     \item 
     % We have successfully constructed a fully quantized spike-driven Transformer (QSD-Transformer) baseline and conducted an in-depth analysis of its performance. Through extensive empirical analysis, we have elucidated the issue of spike information distortion in quantized spike-driven self-attention (Q-SDSA) during the quantization process.
    We construct a lightweight spike-driven Transformer baseline through quantization, called QSD-Transformer. The QSD-Transformer quantizes the synaptic weights from a 32-bit to a low-bit representation (typically 2, 3, and, 4 bits), leading to reduced model size and significant energy efficiency advantages. 
    \item 
    % We propose an information-enhanced spiking neuron that generates multi-bit spikes during training while maintaining the spike-driven nature by extending the virtual timestep during inference and a fine-grained distillation scheme that optimizes the difference between QSD-Transformer and the same architecture ANN attention distribution.
    We reveal that the proposed baseline suffers from performance degradation due to the SID problem in Q-SDSA. 
    Inspired by information entropy, we propose a bi-level optimization strategy to solve this issue.
    This strategy introduces an information-enhanced LIF and a fine-grained distillation to rectify the distribution of Q-SDSA, leading to enhanced performance.
    \item 
    % Our QSD-Transformer represents a pioneering advancement towards accurate and low-bit SD-Transformer models. Experimental results demonstrate that the QSD-Transformer outperforms baseline and existing Spiking Vision Transformer models by a significant margin while retaining an exceptionally low parameter count across various datasets.
    We validate the QSD-Transformer on various visual tasks, e.g., classification, object detection, semantic segmentation, and transfer learning. Experimental results indicate that our method outperforms existing spiking Vision Transformers by a substantial margin, while also boasting a compact model size and extremely low power consumption.
\end{itemize}
%We propose an information-enhanced Spiking Neuron that combines multi-bit values training with spike-driven inference and a Fine-Grained Distillation (FGD) scheme that optimizes the difference between QSD-Transformer and the same architecture ANN attention distribution.
%This spiking neuron activates multi-bit values during training while maintaining the spike-driven nature by extending virtual timestep during inference.
%In the backpropagation phase, and 
% \begin{figure}[!t]
%     \centering
%     \includegraphics[width=0.5\textwidth]{bubble.pdf}
%     \caption{Accuracy vs. Power vs. Params. Our QSD-Transformer enjoys lower power and storage savings while surpassing the best results.}
%     \label{fig:speed-size-acc}
% \end{figure}

\section{Related works}
\noindent \textbf{Spiking vision transformer.}
Spikformer \citep{zhou2023spikformer} pioneered direct training with a pure SNN architecture, introducing a linear self-attention mechanism that eliminates multiplication by activating Query, Key, and Value with spiking neurons and replacing softmax with spiking neurons. Its successor \citep{zhou2024spikformer} integrated masked image modeling \citep{he2022masked}, achieving an 82.25\% accuracy on ImageNet with 172 M parameters, the highest among SNNs.  SpikingResformer \citep{shi2024spikingresformer} introduces a novel spike self-attention mechanism along with a judicious scaling approach, enabling effective extraction of local features. However, none of these models preserved the spike-driven nature until the spike-driven Transformer \citep{yao2023spike}, which introduces the sparse addition to self-attention using only masking operations.   Its successor  \citep{yao2024spike} focused on the meta-design of the spiking vision Transformer, including architecture, spike-driven self-attention, shortcut connections, etc. 
The proposed spike-driven Transformer v2 \citep{yao2024spike} set up direct training SNN backbone for improving performance across tasks like image classification, segmentation, and object detection, hinting at impacts on neuromorphic chip design.  
Hence, in this study, we adopt the pure addition spike-driven Transformer v2 for quantization baseline.
\par
\noindent \textbf{Model compression.}
Numerous compression techniques have been explored to compress large-scale SNNs, including: (1) Pruning \citep{han2015learning,kusupati2020soft,savarese2020winning} in SNNs generally draw on traditional pruning methods from ANNs to suit the spatial and temporal domains \citep{chen2022state,shi2023towards,shen2024efficient}. While successful on simpler datasets and shallow networks, achieving high performance becomes more challenging with complex datasets and deeper networks. (2) Knowledge distillation \citep{hinton2015distilling,guo2023joint,touvron2021training} involves the transfer of knowledge from large-scale ANNs or SNNs to smaller-scale SNNs, aiming to compress models and reduce energy consumption. However, these methods \citep{takuya2021training,tran2022training,xu2023constructing} often distill only the final output of the model, leading to incomplete knowledge transfer and suboptimal performance in SNNs. 
(3) Quantization \citep{jacob2018quantization,krishnamoorthi2018quantizing}, particularly for hardware deployment, is advantageous as it reduces the bit-width of network parameters and activations, enabling efficient inference.
Recent research on quantization methods \citep{stromatias2015robustness,deng2021compress,kheradpisheh2022bs4nn} for SNNs has predominantly focused on weight binarization within Conv-based architectures. For instance, Deng et al. \citep{deng2021compress} utilized QAT \citep{krishnamoorthi2018quantizing,jacob2018quantization} to reduce the weight size of Conv-based SNN, which demonstrated high compression performance with acceptable accuracy loss on recognition tasks. Despite the significant potential of QAT in reducing the memory and computational costs of Conv-based SNN \citep{deng2021compress}, directly applying QAT on Spikformer leads to poor performance. 
The core challenge is the significant distribution discrepancy between the binary spike patterns and the normal distribution in the self-attention of Spikformer and ANN Transformer, which causes severe information distortion, leading to performance degradation.

% of Spikformer and the same architecture ANN Transformer. This discrepancy typically causes a severe information distortion, thereby leading to performance . 
% In Section \ref{method}, we offer a detailed analysis of this challenge and investigate a fine-grained quantization strategy to mitigate this challenge.
% quantification
% To mitigate the impact of quantization on the spiking self-attention mechanism, we propose two fine-grained information compensation modules that improve the robustness of the model against quantization effects.
% Several compression techniques have been explored to address the
% challenges associated with large-scale SNNs. 这些技术包括(1)修剪、(2)知识蒸馏和(3)量化。修剪通过消除网络内部的不必要节点或分支来去除冗余参数。最近SNNs的修剪工作可以大致分为两类。一种将深度神经网络（DNNs）中已建立的修剪方法应用于SNNs的空间和时间域。另一种则使用生物启发式的修剪算法模拟突触再生过程以实现网络压缩。这些方法已成功应用于简单的数据集和浅层网络，但随着数据集和网络变得越来越大和复杂，实现高性能将变得更具挑战性。Knowledge distillation explores transferring knowledge from large-scale ANNs  or SNNs 
% to smaller-scale SNNs to compress the models and reduce energy consumption.然而，这些方法只针对模型的最终输出进行蒸馏，导致SNN没有学到完善的知识，导致性能不佳。其中量化特别适用于在硬件上部署，因为它可以减少网络参数和激活的位宽，以实现高效的推理。Recent research on quantization methods for SNNs mainly focuses on weight binarization of Conv-based architecture. 其中Deng 等人 \citep{deng2021compress}利用量化感知训练(QAT) \citep{krishnamoorthi2018quantizing,jacob2018quantization} 减少卷积架构SNNs的权重大小。他们在基于SNNs的模式识别任务中展示了高压缩性能，同时保持了可接受的精度损失。尽管QAT已经展现出了减少卷积架构SNN模型 \citep{deng2021compress}内存和计算成本的巨大潜力，但对脉冲Transformer模型进行低比特量化的研究仍待深入研究。 % 其根本原因在于脉冲Transformer模型的自注意力机制的前传的二进制脉冲使得模型的准确性对量化非常敏感，这需要更细致的压缩方法。我们提出了两个细颗粒度的信息弥补模块改善了量化对脉冲自注意力机制的影响。

\par
\section{Preliminary}
% In this section, we first introduce the spiking neuron model. Given the absence of prior relevant work, we establish a baseline to evaluate the effects of different module quantization on both the low-bit spike-driven Transformer (QSD-Transformer) \citep{yao2024spike} and its counterpart ANN Transformer \citep{yu2022metaformer}.
In this section, we first introduce the spiking neuron model. Then, we construct a quantized spike-driven Transformer (QSD-Transformer) baseline, which quantizes the synaptic weight from 32-bit to low bit-width, thereby demonstrating significant energy efficiency advantages.
\par
\noindent \textbf{Spiking neuron model.} 
% We integrate the Leaky Integrate-and-Fire (LIF) model into an iterative framework using the Euler method \citep{wu2018spatio} and it can be described as an explicitly iterable version with soft reset:
In this paper, we choose the widely-employed iterative Leaky Integrate-and-Fire (LIF) model \citep{wu2018spatio,guo2024ternary}, which can be described by the following mathematical equations:
\begin{align}
\label{eq:mem} \mathbf{v}^{\ell}[t]&=\mathbf{h}^{\ell}[t-1]+f({\mathbf{w}^{\ell}},\mathbf{x}^{\ell-1}[t-1]), \\
\label{eq:lif}
\mathbf{s}^{\ell}[t]&=\mathbf{\Theta}(\mathbf{v}^{\ell}[t]-\vartheta),\\
\label{eq:headisde} \mathbf{h}^{\ell}[t]&= \tau\mathbf{v}^{\ell}[t]- \mathbf{s}^{\ell}[t],
\end{align}
where $\tau$ is the time constant, $t$ is the time step, $\mathbf{w}^{\ell}$ is the weight matrix of layer $\ell$, $f(\cdot) $ is the operation that stands for convolution (Conv) or fully connected (FC), $\bf x$ is the input, and  $\mathbf{\Theta(\cdot)}$ denotes the Heaviside step function. When the membrane potential $\mathbf{v}$ exceeds the firing threshold $\vartheta$, the LIF neuron will trigger a spike $\mathbf{s}$; otherwise, it remains inactive. After spike emission, the neuron invokes the reset mechanism, where the soft reset function is employed. $\bf h$ is the membrane potential following the reset function. 
% For the backpropagation of this neuron, we outline it in Appendix~\ref{bp}.
\par
\noindent \textbf{QSD-Transformer baseline.} 
We select the purely spike-driven Transformer v2 (SD-Transformer v2) \citep{yao2024spike} to perform quantization, and the LSQ \citep{bhalgat2020lsq+} method is employed to quantize the 32-bit weights to low bit-width (e.g., 2, 3, 4 bits). The quantization function is defined as:
\begin{equation}\label{eq:quantization}
    \begin{aligned}
     % \mathcal{Q}_{\mathbf{w}}({\bf w}) = \lfloor \operatorname{clip}\{\frac{{\bf w}}{\alpha_{\bf w}}, - \mathcal{Q}_n^{\bf w},  \mathcal{Q}_p^{\bf w}\} \rceil, \quad \hat{\bf w} =  \alpha_{\bf w} \mathcal{Q}_{\bf w}({\bf w}),\\
     \mathcal{Q}({\bf w}) = \left \lfloor \operatorname{clip}\left \{ \frac{{\bf w}}{\alpha_{\bf w}}, -2^{b-1},  2^{b-1}\!-\!1 \right \}\right \rceil, \quad \hat{\bf w} =  \alpha_{\bf w} \mathcal{Q}({\bf w}),
    \end{aligned}
\end{equation}
where $\mathbf{w}$ is the 32-bit weight, \(b\) is the bit assigned to the quantized weight (i.e., $\mathcal{Q}({\bf w})$), and \(\alpha_{\mathbf{w}}\) is the scaling factor used to mitigate the quantization error.
Moreover, \(\operatorname{clip}\{x, a, b\}\) confines \(x\) within range \([ a, b]\), and \(\lfloor \cdot \rceil\) denotes the rounding operator.
These two operations make the quantization function non-differentiable, so we adopt the straight-through estimator (STE) \citep{bengio2013estimating} to assist the gradient backpropagation.
Eq.~\ref{eq:quantization} is performed for all weight layers in the baseline model.
Building upon this, the calculation for a certain layer $\ell \in \{\text{FC},\text{Conv}\}$ in our baseline is expressed as:
\begin{equation}\label{eq:act}
    \begin{aligned}
    \mathcal{Q}_{\ell}(\mathbf{x}) = \hat{\bf w}\cdot \mathcal{SN}(\mathbf{x})= \alpha_{\bf w} \mathcal{Q}({\bf w})\cdot \mathcal{SN}(\mathbf{x}).
    \end{aligned}
\end{equation}
Here, $\mathcal{SN}(\cdot)$ represents the spiking neuron layer, which converts the floating-point input $\bf x$ into the binary spike.
Hence, the QSD-Transformer employs binary spike activities and low bit-width weights for Conv and FC operations.
This replaces the original computationally intensive operations, leading to significant energy efficiency improvements. Following Eq. \ref{eq:act}, the quantization for the spike-driven self-attention (Q-SDSA) can be further described as: 
\begin{equation}
    \mathbf{q}=\mathcal{Q}_{\text{Conv1}}(\mathbf{x}),\mathbf{k}=\mathcal{Q}_{\text{Conv2}}(\mathbf{x}),\mathbf{v}=\mathcal{Q}_{\text{Conv3}}(\mathbf{x}), \quad \text{Q-SDSA}(\mathbf{q},\mathbf{k},\mathbf{v})= \mathcal{SN}((\mathbf{q}_{\bf s} \mathbf{k}_{\mathbf{s}}^{T}) \mathbf{v}_{\bf s}),
    \label{eq:att}
\end{equation}
where, $\bf{q_s}=\mathcal{SN}(\bf q)$, and $\bf{k_s}$, $\bf{v_s}$ are obtained in the same way.
It can be observed from Eq.~\ref{eq:att} that our Q-SDSA module reduces the computational number by the linear attention mechanism with $\mathcal{O}(N D^{2})$, where $N$ is the token numbers and $D$ is the channel dimensions.
However, quantizing the attention module will diminish its representation capacity, leading to performance degradation.
In the next section, we will provide a detailed explanation of this issue and propose methods to address it.

% For the same architecture ANN Transformer, we replace the original SDSA  with vanilla self-attention while modifying the Conv and FC layers' spiking neurons to StarReLU according to Conv-Attention Transformer (CAformer) \citep{yu2022metaformer}. And we also employ LSQ \citep{bhalgat2020lsq+} for weight and activation quantization.

% $\mathcal{SN}(\cdot)$ denotes spiking neurons, and $\bf x$ generally denotes the activation in this paper, including the input feature map of FC and Conv layers and the input of self-attention modules. 
% $\ell \in \{\text{FC},\text{Conv}\}$ is the FC or Conv layer. Moreover, the straight-through estimator (STE) \citep{bengio2013estimating} is used to retain the derivation of the gradient in backward propagation of weight $\bf w$.
% The SD-Transformer is a hybrid CNN-VIT framework. It employs multiple convolutional layers at the front to enhance inductive bias and the quantized framework can be described by Eq \ref{eq:act}. Following this, the quantized spike-driven self-attention (Q-SDSA) mechanism \citep{yao2024spike} can be outlined as follows:

\section{Method}
\label{method}
% In this section, we first find the QSD-Transformer baseline model exhibits a significant performance gap compared to the full-precision model. Our research indicates that this baseline model suffers from severe information distortion in the quantized attention scores during forward propagation. To address this, we formulate it as a bilevel optimization problem. In the lower level, we use a new spiking neuron to align the distribution of attention scores, maximizing self-information entropy. At the upper level, we introduce a grained distillation scheme, which effectively transfers teacher information into the distilled features, thereby minimizing conditional information entropy.

In this section, we first reveal that the performance degradation of our baseline is due to the limited representational capacity of Q-SDSA. Inspired by information entropy theory, we propose a bi-level optimization strategy to address this issue. 
At the lower level, we introduce the information-enhanced leaky integrate-and-fire (IE-LIF) neuron, which maximizes information entropy by adjusting the spike distribution. At the upper level, we present the fine-grained distillation scheme, which minimizes conditional entropy by aligning the information of Q-SDSA with that of ANNs.

\begin{figure}[ht]
\centering
\subfigure[]{\includegraphics[width=0.37\linewidth]{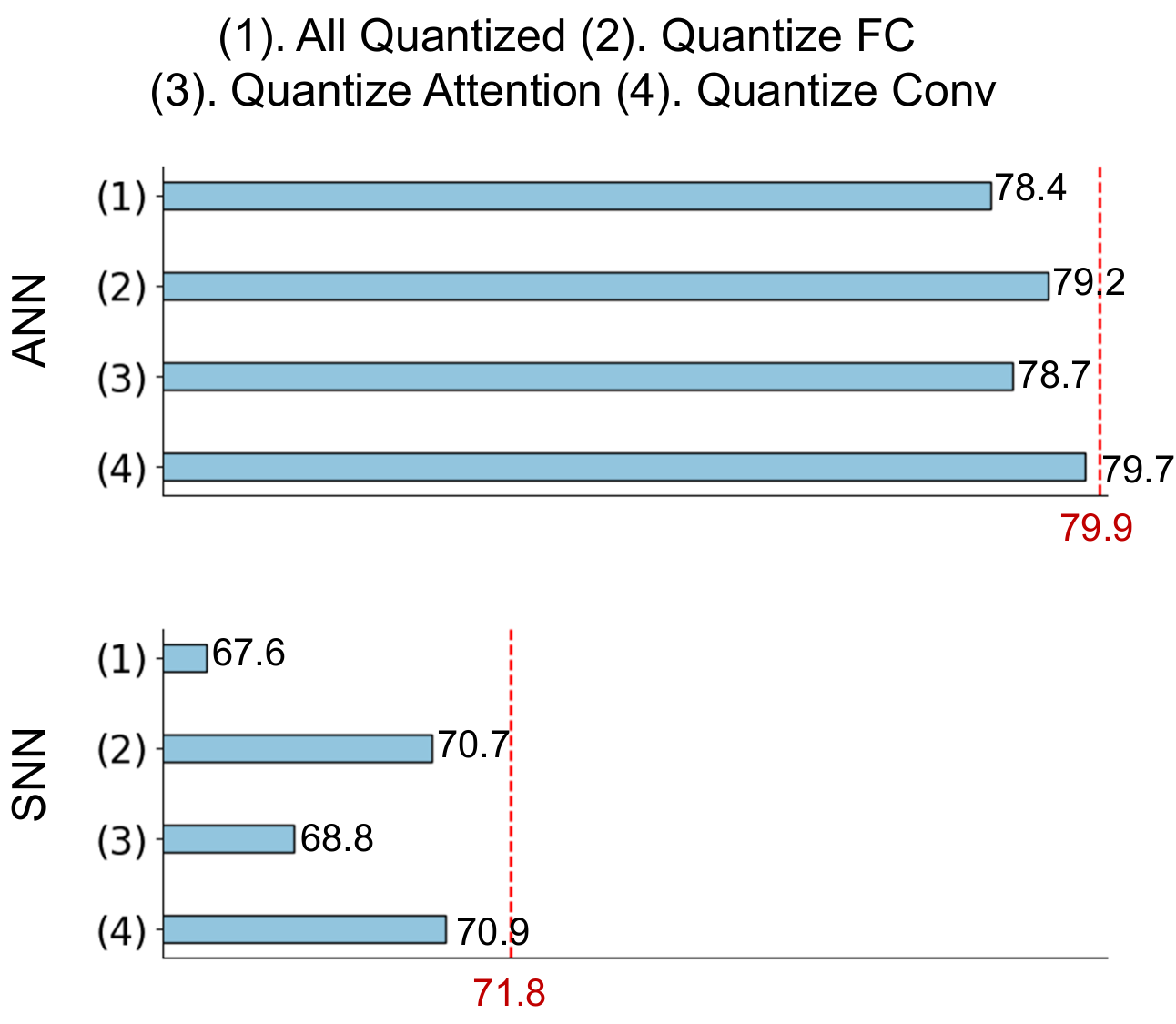}}
\quad 
\subfigure[]{\includegraphics[width=0.57\linewidth]{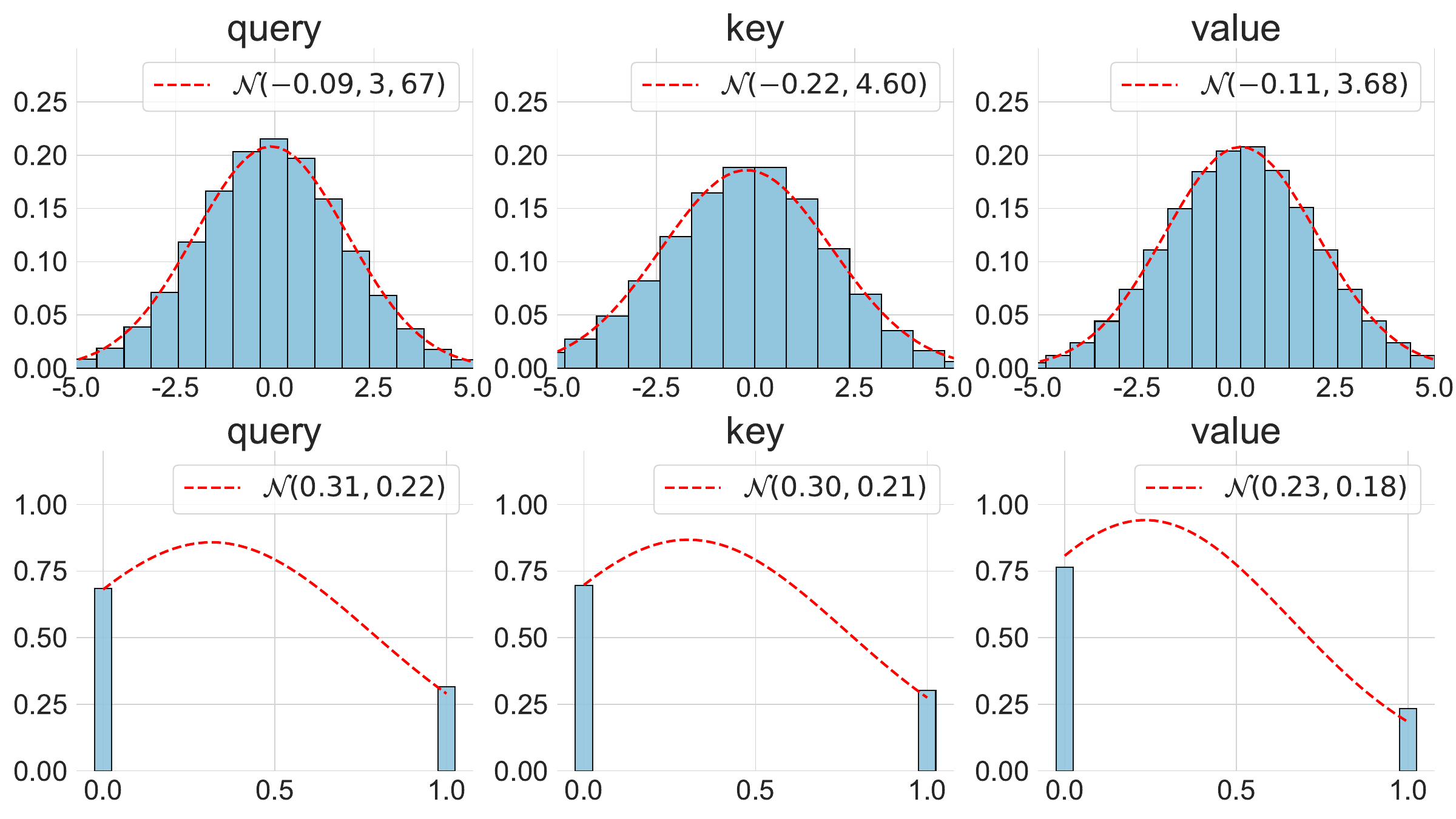}}
\caption{(a) Accuracy of quantizing different modules in the SD-Transformer v2 and its same ANN Transformer. (b) The distribution of the attention module (blue shadow), and the probability density function curve of normal distribution (red line). Experiments are conducted on ImageNet, and 3 layers in SD-Transformer v2 and its same ANN Transformer are selected for illustration.}
\label{fig:degrad}
\end{figure}
\subsection{Performance degradation analysis}
\label{degrad:sec}
Despite its efficiency advantages, the QSD-Transformer baseline suffers from performance degradation, as shown in Fig.~\ref{fig:degrad} (a) (All quantized).
In contrast, the quantized ANN Transformer can balance both efficiency and performance \citep{wu2022tinyvit}.
Hence, we quantize each module of the architecture, i.e., FC, Conv, and Attention, to identify which one has the biggest impact on performance degradation.
\par
We illustrate the ablation results in Fig.~\ref{fig:degrad} (a), where all weights are quantized to 4-bit. Obviously, the attention layer in SNNs is highly sensitive to quantization, but it is not observed in ANNs.
This is attributed to the different information distributions in the self-attention map, as depicted in Fig.~\ref{fig:degrad} (b).
It can be observed that the information within Q-SDSA displays a bimodal distribution, whereas the information in ANN adheres to a normal distribution. 
Due to the utilization of both low-bit weights and binary spikes, the information representation capability of the Q-SDSA is severely limited compared to that of ANNs. We define it as the spike information distortion (SID) problem.

% quantizing the weights of the spike-driven Transformer \citep{yao2024spike} to 4 bits results in a performance drop of 4.2\%, with the most significant impact observed in the quantization of the SDSA module, which accounts for a 3.0\% drop. In contrast, quantizing the weights and activation values of the same ANN architecture \citep{yu2023metaformer} to 4 bits results in only a 1.5\% performance degradation. 
% It can be found that the attention layer in SNNs is highly sensitive to quantization, but it is not observed in ANNs.

% We attribute this discrepancy to the mismatch between the distribution of Q-SDSA in the QSD-Transformer compared to the Multi-head Self-attention (MHSA) in the same ANN Transformer architecture. The self-attention mechanism in Transformers is designed to model long-range dependencies. For instance, as shown in Fig. \ref{fig:degrad} (b), the distribution of the first layer query in the ANN approximates a normal distribution $\mathcal{N}(\mu,\sigma)$, whereas in the SNN, it exhibits an extreme binary distribution. Consequently, the use of low-bit Q-SDSA inevitably reduces the attention module's capability to capture global dependencies in the input.  This discrepancy is exacerbated as the model size increases, leading to a sharp decline in performance.

% \subsection{\textcolor{red}{Information Content Analysis of QSD-Transformer}}
To solve the SID issue, we draw on the quantized Transformer in the ANN domain \citep{liuptq} that has struck a good balance between efficiency and performance by maintaining the noraml distribution of activity. This prompts us to adjust the information distribution in the Q-SDSA to match that of ANNs.
To achieve this, we take inspiration from the information entropy theory \citep{paninski2003estimation} and formulate it as the mutual information entropy maximization problem.
% We summarize the above phenomenon as Spike Information Distortion (SID). Then we delve into the information entropy and knowledge distillation and scrutinize its potential to mitigate the SID phenomenon and enhance the performance of QSD-Transformer.
% The ANN Transformer has demonstrated exceptional performance under low-bit compression, prompting us to employ a full-precision ANN Transformer architecture as the teacher and a Quantized SNN Transformer (QSD-Transformer) as the student, indicated by the $\mathcal{A}$ and $\mathcal{S}$, respectively. \par
\par
\noindent \textbf{Definition 1.}  \textit{Addressing the performance degradation of the QSD-Transformer baseline is equivalent to maximizing the mutual information entropy between it and the quantized Transformer in ANNs. The optimization goal for the QSD-Transformer is defined below.}
\begin{equation}
\label{eq:mutual}
    \mathop{\max}_{\theta^{\mathcal{S}}} \mathcal{I}({\bf p}^{\mathcal{S}}; {\bf p}^{\mathcal{A}}) =\mathcal{H}({\bf p}^{\mathcal{S}}) - \mathcal{H}({\bf p}^{\mathcal{S}}|{\bf p}^{\mathcal{A}}),
\end{equation}
where ${\bf p}^{\mathcal{S}}$ and ${\bf p}^{\mathcal{A}}$ are the attention score value in SNN and ANN respectively, and $\theta^{\mathcal{S}}$ is the parameters of the QSD-Transformer.
However, directly optimizing Eq.~\ref{eq:mutual} is challenging, so we regard it as a bi-level optimization problem \citep{colson2007overview,sinha2017review}.
It is achieved by minimizing the conditional information entropy $\mathcal{H}({\bf p}^{\mathcal{S}}|{\bf p}^{\mathcal{A}})$ and maximizing the information entropy $\mathcal{H}({\bf p}^{\mathcal{S}})$, which is defined as:
% According to \textbf{Definition 1}, we can enhance performance by minimizing the mutual information entropy between the QSD-Transformer and its same architecture ANN while maximizing the QSD-Transformer's information entropy. However, simultaneously optimizing this goal is challenging. Thus, we reformulate Eq. \ref{eq:mutual} as a bi-level problem \citep{colson2007overview,sinha2017review}, alternately optimizing these terms, which is defined as follows:
% 根据\textbf{定义1.}，我们可以通过最小化学生和老师互信息熵和最大化学生的信息熵来提升性能，然而，同时优化上述最大和最小项是具有挑战性的。 相反，我们做出妥协，重新表述方程。 \ref{eq:mutual} 作为一个双层问题 \citep{colson2007overview,sinha2017review} 交替优化这两个项目。并将其重新定义为：
\begin{equation}
\mathop{\min}_{\theta^{\mathcal{S}}} \mathcal{H}({\bf p}^{\mathcal{S}^\star}|{\bf p}^{\mathcal{A}}), \quad
\operatorname{s.t.} \quad {\bf p}^{\mathcal{S}^\star} = \mathop{\arg \max}_{{\bf p}^{\mathcal{S}}} \mathcal{H}({\bf p}^{\mathcal{S}}).
\label{eq:bi-level}
\end{equation}
To accomplish it, we first propose the information-enhanced LIF (IE-LIF) neuron, aiming to maximize the information entropy ${\bf p}^{\mathcal{S}^{\star}}$ at the lower level.
%虽然对于SNN来说通过最大化脉冲传输信息熵提升性能已经十分常见，但如何保证
We further introduce a fine-grained distillation (FGD) scheme, aiming to minimize the conditional entropy $\mathcal{H}({\bf p}^{\mathcal{S}}|{\bf p}^{\mathcal{A}})$ at the upper level.

% derive the highest information entropy ${\bf p}^{\mathcal{S}^{\star}}$  at the lower-level. 
% Here, the goal involves two subproblems. We first solve the lower-level problem by using proposed information-enhanced LIF (IE-LIF) forward-passing to maximize the information entropy $\mathcal{H}({\bf p}^{\mathcal{S}})$, deriving the current highest information entropy ${\bf p}^{\mathcal{S}^{\star}}$. This is then used in the upper-level problem during backpropagation by proposed fine-grained distillation (FGD) to minimize the conditional entropy $\mathcal{H}({\bf p}^{\mathcal{S}}|{\bf p}^{\mathcal{A}})$ until the loss function converges.
%这样的目标涉及两个子问题，我们先求解下层问题，%在每次训练迭代中，向前求解最大化自信息熵 $H({\bf q}^{\mathcal{S}})$，并导出当前信息熵最大的 ${\bf p}^{\mathcal{S}^{\star}}$ 送给上层在反向传播时最小化条件熵 $H({\bf q}^{\mathcal {S}}|{\bf q}^{\mathcal{T}})$，进行迭代直到损失函数收敛。

%在每次训练迭代中，向前求解最大化自信息熵 $H({\bf q}^{\mathcal{S}})$，同时最小化条件熵 $H({\bf q}^{\mathcal {S}}|{\bf q}^{\mathcal{T}})$是在后向过程中实现的。

%为此我们探索信息熵的概念，并分析如何利用它来解决spike information distortion提高QSD-Transformer的性能。
%ANN Transformer在低比特压缩上表现出了卓越的性能，因此为了解决QSD-Transformer的信息失真问题，我们旨在知识蒸馏框架下提高量化网络的表示能力。
%我们的QSD-Transformer追求性能和压缩之间的最佳平衡，这正是通过量化中间层关于输入（越少越好）和期望输出（越多越好）的互信息的信息瓶颈（IB）方法的目标~\citep{shwartz2017opening,tishby2000information}。
%我们的QSD-Transformer追求性能和压缩之间的最佳平衡，这是通过提升我们架构中Q-SDSA中和同架构ANN Transformer的query,key,value和的相似度来实现的.
%因此，我们QSD-Transformer的目标对象是：

%因此，我们QSD-Transformer的目标对象是：：

\subsection{Information-enhanced LIF neuron}
\label{iesn}
%在进行\noindent \textbf{Low-level optimization.}之前，我们
%将膜电位量化为二元尖峰将导致 Q-SDSA 中查询的二元分布，与 ANN Transformer 中的正态分布查询相比，减少信息丢失和表示能力，如图所示。 \ref{fig:degrad} ( b)，发现query二项分布不可能去拟合正态 idismatch 很多。为此在进行进行\noindent \textbf{Low-level optimization.}之前，我们先构造了一个IE-LIF模型。
% Quantizing the membrane potential into binary spikes results in binomial distributions of attention scores in Q-SDSA, leading to significant information loss and reduced representational capacity compared to the normally distributed attention scores in the ANN Transformer, as illustrated in Fig. \ref{fig:degrad} (b). To mitigate this issue, we initially developed an information-enhanced LIF (IE-LIF) module prior to undertaking low-level optimization.

As mentioned in Fig. \ref{fig:degrad} (b), the information in the self-attention map of the QSD-Transformer follows a binomial distribution, which limits the representational capacity of the attention module. Therefore, we propose the information-enhanced LIF (IE-LIF) neuron and adjust the information distribution of Q-SDSA at the lower level, focusing on maximizing the information entropy $\mathcal{H}({\bf p}^{\mathcal{S}})$.

% Quantizing the membrane potential into binary spikes results in binomial distributions of attention scores in Q-SDSA, reducing information loss and representation compared to the normally distributed attention scores in the ANN Transformer, as shown in Fig. \ref{fig:degrad} (b). To address this, we first constructed an information-enhanced LIF (IE-LIF) module before performing low-level optimization.
\par
%假设SNN Transformer和ANN Transformer中的q,k,v分布服从二值和正态分布，当SNN中的时间步长趋近于无穷时，存在一组超参数$\mu,\sigma,r$使得，$\mathcal{H}(\mathbf{p})$=$\mathcal{H}(\mathbf{p})$, where $p \in \{q,k,v\}.$
\par
\noindent \textbf{Proposition 1.} \textit{
Given the SNN Transformer and ANN Transformer models, where the distributions of the query ($\mathbf{q}$), key ($\mathbf{k}$), and value ($\mathbf{v}$) follow binomial $\mathcal{B}(r, T)$ and normal $\mathcal{N}(\mu,\sigma)$ distributions, respectively, it is postulated that as the SNN's time step $T$ tends to infinity, there exist parameters $\mu$, $\sigma$, and $r$ such that the average entropy over time of the SNN's attention scores $\mathcal{H}(\sum_{t=1}^{T}\mathbf{p}^\mathcal{S}[t])$ equals ANN attention scores' entropy $\mathcal{H}(\mathbf{p}^\mathcal{A})$.}
% \par
% \noindent \textbf{Proposition 1.} \textit{Given the SNN Transformer and ANN Transformer models, where the distributions of the query ($\mathbf{q}$), key ($\mathbf{k}$), and value ($\mathbf{v}$) follow binomial $\mathcal{B}(r, T)$ and normal $\mathcal{N}(\mu,\sigma)$ distributions, respectively. As the SNN's time step $T$ tends to infinity, $\exists \mu,\sigma, r $ such that $\mathcal{H}(\sum_{t=1}^{T}\mathbf{p}^\mathcal{S}[t])=\mathcal{H}(\mathbf{p}^\mathcal{T})$.}
% where $\mathbf{p}_\mathbf{s}[t] \in \{\mathbf{q}_\mathbf{s}, \mathbf{k}_\mathbf{s} \mathbf{v}_\mathbf{s}\}$ and $ \mathbf{p}_\mathbf{a} \in \{\mathbf{q}_{fp}, \mathbf{k}_{fp}, \mathbf{v}_{fp}\}.$}
\par
Proof can be found in Appendix \ref{lim}. According to Proposition 1., within infinite time steps $T$, the attention matrix values (e.g., $\mathbf{q}_\mathbf{s}, \mathbf{k}_\mathbf{s} \mathbf{v}_\mathbf{s}\}$) in the QSD-Transformer have the same information representation as those in the ANN Transformer.
However, numerous time steps $T$ will inevitably lead to latency and  huge energy consumption.
%然而大量的时间步长will boost高延迟和功耗
Recently, \cite{hao2023bridging} and \cite{hu2023fast}  achieved high-performance conversion by transforming quantized ANNs into SNNs and using fewer time steps. This prompted us to train directly using multi-bit values.
%根据 \textbf{Proposition 1.}, our QSD-Transformer在T无穷大时与ANN Transformer的注意力矩阵值如query表达能力相同，but this requires a long simulation time step and boosts the energy consumption. 
%提高了q,k,v的分布的方差，增加了信息熵$$,近似为离散的正态分布
We only need to ensure that inference is spike-driven.  Thus, we propose the concept of IE-LIF, in which Eq. \ref{eq:lif} can be written as:
\begin{equation}
    \mathbf{a}^{\ell}[t]=\frac{1}{b} \left \lfloor  \text{clip}\{
    \mathbf{v}^{\ell}[t],0,b\} \right \rceil,
    \label{new}
\end{equation}
where $ \mathbf{a}^{\ell}[t]$ is the multi-bit output of IE-LIF, and $b$ represents the maximum integer value emitted by the IE-LIF, which is equipped with the baseline.
% , which allowed the query in the attention matrix to approximate a discrete normal distribution and improve information entropy. 
Since Eq. \ref{new} is non-differentiable, we employ the straight-through estimator (STE) \citep{bengio2013estimating} to retain the gradient derivation during backpropagation. 
\par
Previous SNNs have utilized multi-bit spikes (integers) \citep{hao2024lm,ponghiran2022spiking} or continuous values \citep{wu2021liaf} to reduce quantization error, thereby alleviating the shortcomings of binary spikes. However, this approach raises concerns because it can undermine the inherent spike-driven characteristics of SNNs. We propose a solution where IE-LIF uses multi-bit values during the training phase and subsequently converts these values to binary spikes for inference, as depicted in Fig. \ref{fig:method} (c).
Moreover, the output $\mathbf{x}^{\ell+1}[t]$ of each layer in the SNN is represented as:
\begin{equation}
\mathbf{x}^{\ell+1}[t] = \mathbf{w}^{\ell} \cdot  \mathbf{a}^{\ell}[t] = \mathbf{w}^{\ell} \cdot \sum_{t=1}^{T} \mathbf{s}^{\ell}[t],
\end{equation}
where $a^{\ell}[t]$ represents multi-bit spikes during training with one timestep and is denoted as $\{0,0.25,0.5,0.75,1\}^{T=1}$, while $s^{\ell}[t]$ represents binary spikes during inference and is extended to 4 virtual timesteps denoted as $\{0,1\}^{T=4}$, with a maximum integer value $b$ set to 4 in this paper.
\par
% 有了IE-LIF之后接下来我们将进行Low-level optimization 去最大化注意力分数的熵
% \par\noindent \textbf{Low-level optimization.} 
With the introduction of IE-LIF, our next step involves low-level optimization to maximize the entropy of attention scores ${\bf p}^{\mathcal{S}}$. We first observed that the membrane potentials $\{\mathbf{q}_{\text{mem}}, \mathbf{k}_{\text{mem}}, \mathbf{v}_{\text{mem}}\}$ in the IE-LIF model within Q-SDSA approximately follow a normal distribution $\mathcal{N}(\mu, \sigma)$ with $\mu = 0$, as also noted in \citep{guo2022loss,guo2022reducing}. Then we provide the formula for calculating the maximum information entropy.
%我们首先发现膜电位 $\{\mathbf{q}_{\text{mem}},\mathbf{k}_{\text{mem}},\mathbf{v}_{\text{mem}}\ Q-SDSA 中 IE-LIF 模型的 $\mathcal{N}(\mu,\sigma)$ 近似服从正态分布 $\mu=0$，这也在\citep{}中被观察到，为此在接下来的理论中我们先给出膜电位的信息熵计算公式，而通过IE-LIF之后的脉冲注意力分数的信息熵可以表示为对膜电位信息熵的离散抽样。
% 然而，膜电位 $\{\mathbf{q}_{\text{mem}},\mathbf{k}_{\text{mem}},\mathbf{v}_{\text{mem}}\ Q-SDSA 中 IE-LIF 模型的 $\mathcal{N}(\mu,\sigma)$ 近似服从正态分布 $\mu=0$。 
% 之后在接下来的理论中我们给出联系变量的信息熵计算
%为了计算通过IE-LIF之后的脉冲注意力分数的信息熵，
%通过IE-LIF之后的脉冲注意力分数的信息熵可以表示为对膜电位信息熵的离散抽样，:

% and is observed to increase as the variance $\sigma$ expands
\par\noindent \textbf{Proposition 2.} \textit{Given a random variable $\mathbf{x}$ following a normal distribution $\mathcal{N}(\mu,\sigma)$, the information entropy $ \mathcal{H}(\mathbf{x})$ achieves its maximum value of $\frac{1}{2}\log 2\pi e \sigma^{2}(\mathbf{x}) $.}
\par
Proof can be found in Appendix \ref{entropy}. According to Proposition 2., the maximum information entropy of membrane potential is $\mathcal{H}(\mathbf{p}_{\text{mem}})=\frac{1}{2}\log 2\pi e \sigma^{2}(\mathbf{p}_{\text{mem}})$, acting as the upper limit for the information entropy of spike attention scores. Through the application of IE-LIF, the information entropy of spike attention scores becomes a discrete representation of membrane potential's information entropy:
\begin{equation}
    \mathcal{H}({\bf p}^{\mathcal{S}}) = -\sum_{k=0}^{b}\left(G({\mathbf{p}}_\text{mem})\delta({\mathbf{p}}_\text{mem}-\frac{k}{b})\right) \cdot
    \log \left(G({\mathbf{p}}_\text{mem})\delta({\mathbf{p}}_\text{mem}-\frac{k}{b})\right),
\end{equation}
where $G(\mathbf{p}_\text{mem})$ is the Gaussian distribution function of  membrane potential $\mathbf{p}_\text{mem}$ and $\delta(\cdot)$ is the Dirac delta function and $\int_{-\infty}^{\infty} \delta(x) \, dx = 1$. However, since $\mathbf{p}_\text{mem} \sim \mathcal{N}(0,\sigma)$, when Eq. \ref{new} is applied to membrane potentials $\mathbf{p}_\text{mem}$, spikes are emitted only when $\mathbf{p}_\text{mem}$ exceeds 0. 
This may lead to the distributions of attention scores resembling the right tail of a discrete normal distribution, causing mismatched attention scores between SNNs and ANNs.
% 这可能会导致查询的分布和离散正态分布的右半部分右边相似，从而导致 SNN 中的注意力分数与 ANN 中的注意力分数不匹配。
% This may cause the distribution of queries to approximate only the right half of a discrete normal distribution, resulting in a mismatch between the attention scores in SNNs and those in ANNs.
 % Hence, we propose a membrane potential rectify function (MPRF) $\phi^{\ell}(\cdot)$ to maximize the information entropy of attention score, which can be expressed as:
 Hence, we propose a membrane potential rectify function (MPRF) $\phi^{\ell}(\cdot)$, which can be expressed as:
\begin{equation}
\hat{\mathbf{p}}_\text{mem}=\phi^{\ell}(\mathbf{p}_{\text{mem}}) = 
\frac{\mathbf{p}_{\text{mem}}-\mu(\mathbf{p}_{\text{mem}})}{\sigma(\mathbf{\mathbf{p}_{\text{mem}}})}\cdot \gamma  +\alpha,
\label{MPRF}
\end{equation}
where $\mathbf{p}_{\text{mem}}$ and $\hat{\mathbf{p}}_\text{mem}$ represent the membrane potential of Q-SDSA before and after applying the MPRF, and $\gamma, \alpha$ are the learnable hyperparameters. The MPRF is only executed when $\ell \in \text{Q-SDSA}$.  After MPRF, attention score distribution aligns more closely with the desired normal distribution, reducing the mismatch between SNNs and ANNs. At this point, the information entropy of SNN attention scores is ${\bf p}^{\mathcal{S}^\star} = \sum_{k=0}^{b}G(\hat{\mathbf{p}}_\text{mem})\delta(\hat{\mathbf{p}}_\text{mem}-\frac{k}{b})$. 
Furthermore, the MPRF can be incorporated into the weights $\bf w^{\ell}$ during inference, details of which can be found in Appendix \ref{Fusion}.
\par
\subsection{Fine-grained distillation}
The IE-LIF neuron has maximized the information entropy of $\mathcal{H}({\bf p}^{\mathcal{S}})$.
Building upon this, we achieve the optimization goal of Eq.~\ref{eq:bi-level} by proposing a fine-grained distillation (FGD), which adjusts the distribution of Q-SDSA at the upper level to minimize the conditional entropy $\mathcal{H}({\bf p}^{\mathcal{S}^\star}|{\bf p}^{\mathcal{A}})$.
% After get the ${\bf p}^{\mathcal{S}^\star} = \mathop{\arg \max}_{{\bf p}^{\mathcal{S}}} \mathcal{H}({\bf p}^{\mathcal{S}})$ by the above IE-LIF, the corrected upper-level optimization by our proposed fine-grained distillation (FGD) in Eq. \ref{eq:bi-level} can be rewritten as: $\mathop{\min}_{\theta^{\mathcal{S}}}\mathcal{H}({\bf p}^{\mathcal{S}^\star}|{\bf p}^{\mathcal{A}})$. 
\par
% However, minimizing the aforementioned formula presents challenges. 
The proposed FGD achieves minimal conditional entropy by minimizing the norm distance between $\hat{\mathbf{p}}^{\mathcal{S}^{\star}}$ and $\hat{\mathbf{p}}^{\mathcal{A}}$, with the optimal solution being $\hat{\mathbf{p}}^{\mathcal{S}^{\star}} = \hat{\mathbf{p}}^{\mathcal{A}}$. It utilizes appropriate distillation activations and meticulously designed similarity matrices to effectively leverage knowledge from the teacher model. 
Therefore, the FGD scheme is defined as:
\begin{equation}
\mathcal{L}_\text{FGD}=\sum_{\mathbf{p \in \{\mathbf{q},\mathbf{k},\mathbf{v}\}}}\sum_{l=1}^{L}\sum_{h=1}^{H}||\mathcal{F}_{\mathbf{p}^{\mathcal{A}}}^{l} - \mathcal{F}_{\mathbf{p}^{\mathcal{S}}}^{l}||, \quad \text{where} \quad \mathcal{F}_{\mathbf{p}} = \frac{\mathbf{p}\times \mathbf{p}^{\top}}{||\mathbf{p}\times\mathbf{p}^{\top}||},
\end{equation}
where $L$ denotes the number of layers in the Transformer, $H$ represents the number of heads, and $||\cdot||$ indicates $\ell_{2}$ normalization. During backpropagation, gradient updation drives the attention matrix of the QSD-Transformer and the same ANN Transformer closer, thereby minimizing $\mathcal{H}(\mathbf{p}^{\mathcal{S}^{\star}}|\mathbf{p}^{\mathcal{A}})$. 
The overall training loss function $\mathcal{L}$ of our QSD-Transformer is defined as:
\begin{equation}
\mathcal{L}=\mathcal{L}_{\text{CE}}(\sum_{t=1}^{T} \mathbf{s}^{\ell}[t],y)+\lambda \mathcal{L}_\text{FGD},
\end{equation}
where $\mathbf{s}^{\ell}[t] $ is the output of our QSD-Transformer, $\mathcal{L}_{\text{CE}}$ is the cross-entropy loss \citep{rathi2021diet} for ensuring task performance, and $\mathcal{L}_{\text{FGD}}$ is the proposed distillation loss for enhancing information entropy. $\lambda$ is a coefficient to balance these two loss functions, and it is set to 2 in this paper.
% We introduce a variant of the distillation method where we treat the teacher's hard decision as the ground truth. Denoting $y_t = \text{argmax}_c Z_t(c)$ as the teacher model's hard decision, the objective function related to distillation with this hard label is formulated as:
% \begin{equation}
%     \mathcal{L}_{dist} = \frac{1}{2} \mathcal{L}_{\text{CE}}(\psi(Z_s), y) + \frac{1}{2} \mathcal{L}_{\text{CE}}(\psi(Z_s), y_t),
% \end{equation}
% where $Z_t$ represents the teacher model's classification results, $Z_s$ represents the student model's classification results, $y$ is the true label, and $\psi$ represents the softmax function.
% \begin{figure}[htbp]
%     \centering
%     \includegraphics[width=1\linewidth]{after.pdf}
%     \caption{(a) Our QSD-Transformer enjoys lower power and storage savings while accuracy surpasses the best results. (b) IE-LIF and FGD can reduce spike information distortion of baseline.}
%     \label{fig:after}
% \end{figure}
\begin{figure}[!t]
\centering
\subfigure[]{\includegraphics[width=0.39\linewidth]{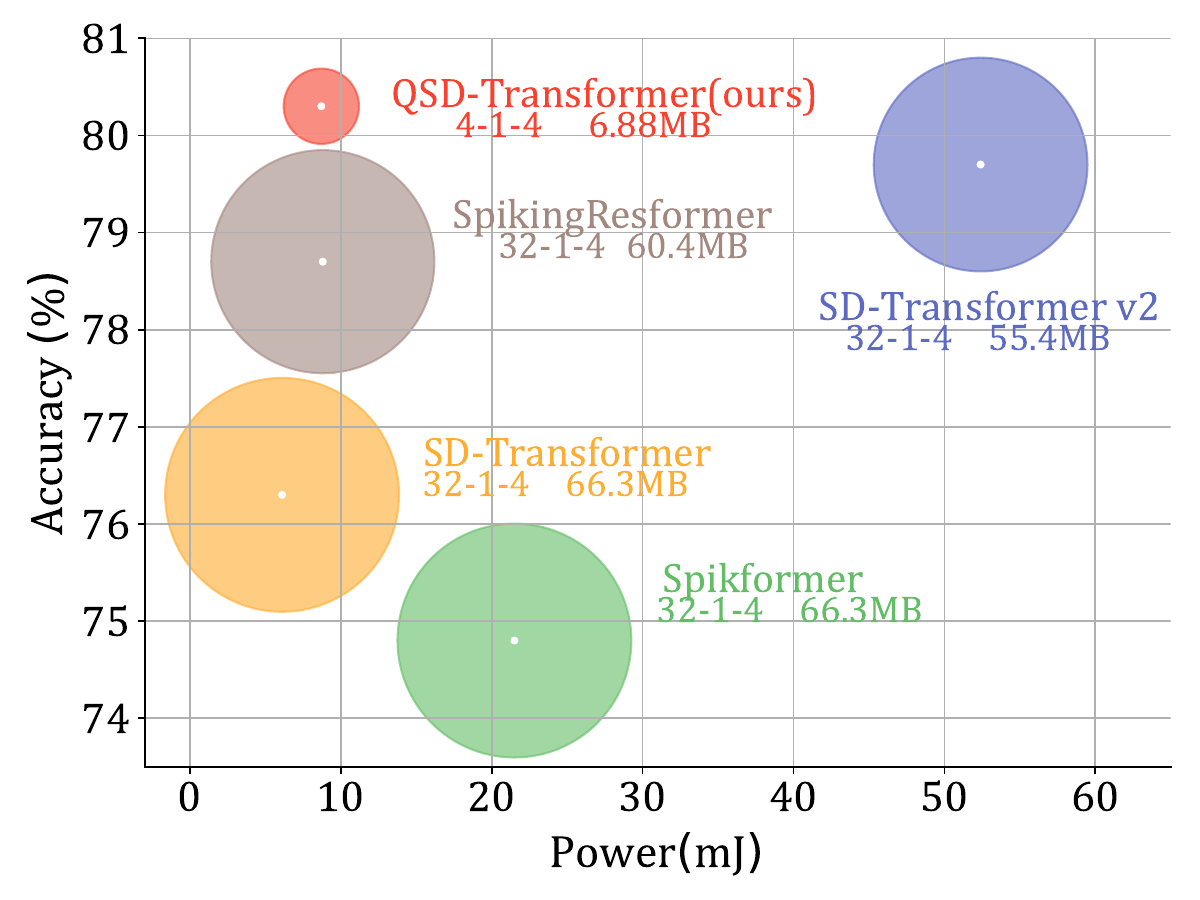}}
\quad 
\subfigure[]{\includegraphics[width=0.54\linewidth]{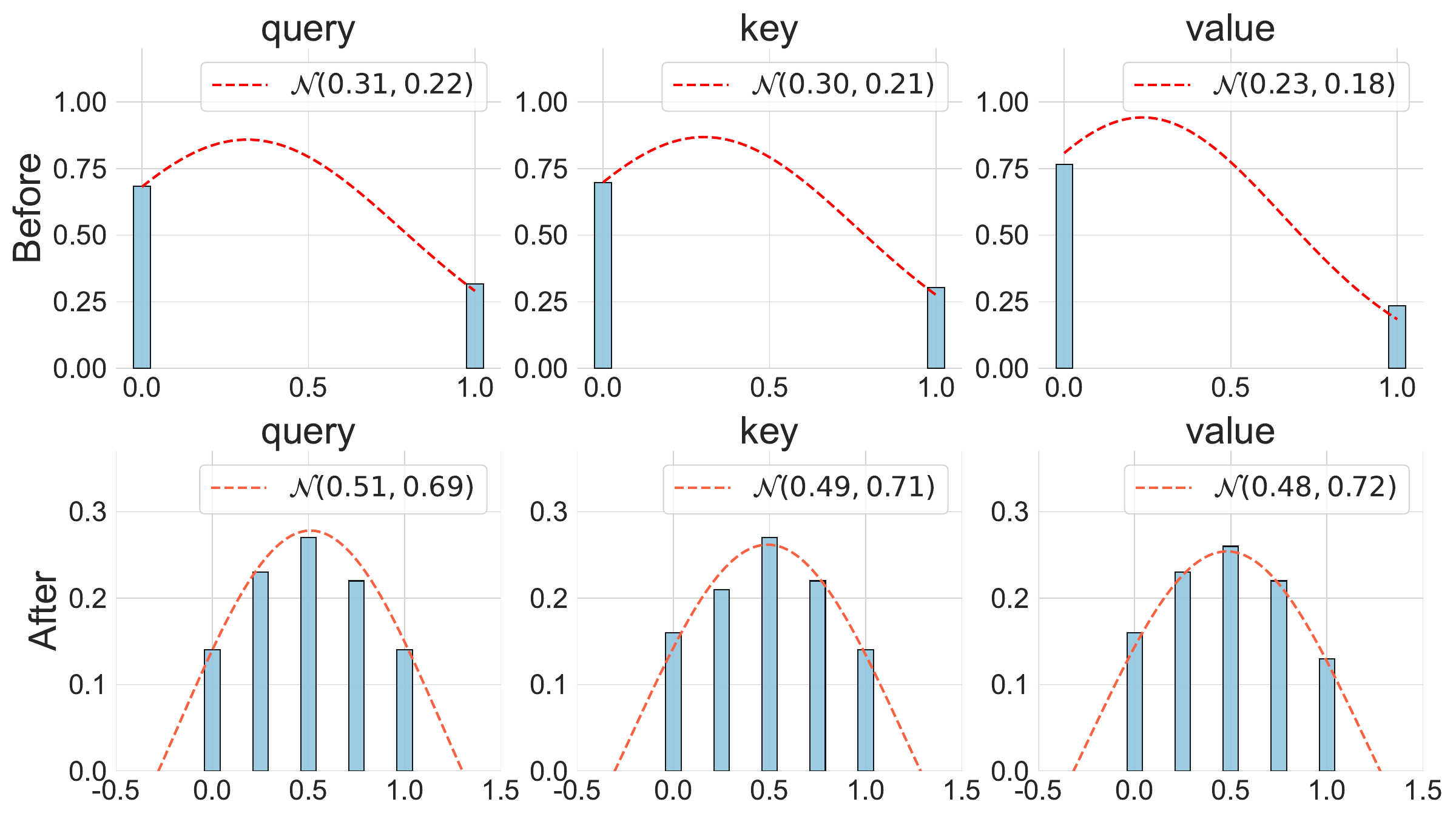}}
\caption{(a) Comparative results of accuracy, power, and parameters on ImageNet. (b) Comparison of information distribution in Q-SDSA before and after using the proposed IE-LIF and FGD scheme.}
\label{fig:after}
\end{figure}
\vspace{-4mm}
\section{Experiment}
In this section, we validate the QSD-Transformer on various vision tasks, including image classification, object detection, semantic segmentation, and transfer learning.
Then, we ablate the proposed  scheme to prove the effectiveness of our method.
For further detailed information on datasets, power calculations, experimental setups, and hyperparameters, refer to Appendix \ref{Energy} and \ref{details}.

% In this section, we initially assess the QSD-Transformer's performance on ImageNet classification \citep{krizhevsky2012imagenet}. Subsequently, we utilize the ImageNet-trained model to fine-tune detection or segmentation heads on COCO2017 \citep{lin2014microsoft} and ADE20K \citep{zhou2017scene} datasets. Additionally, we evaluate the transfer learning capability of the QSD-Transformer across various datasets. Finally, we conduct ablation experiments on IE-LIF, FGD, and weight bits within the QSD-Transformer. For further information on datasets, experimental setups, and hyperparameters, refer to Appendix \ref{details}.
% 我们we perform ablation experiments on xxx and xxx in QSD-Transformer

% %文杰师姐
%     \begin{figure}[!t]
%         \centering
%         \includegraphics[width=0.5\textwidth]{bubble.pdf}
%         \caption{Accuracy vs. Power vs. Params. Our QSD-Transformer enjoys lower power and storage savings while surpassing the best results.}
%         \label{fig:speed-size-acc}
%     \end{figure}
%我们测试了所提出的IE-LIF和FGD对QSD-Transformer中的SDSA

\par\noindent\textbf{ImageNet classification.}
We evaluate the QSD-Transformer's effectiveness in image classification using the challenging ImageNet-1K dataset \citep{deng2009imagenet}. 
% This dataset comprises approximately 1.3M training images and 50K validation images, covering common 1000 classes.
% We conduct a comprehensive comparison with existing advanced deep models, such as Metaformer~\citep{yu2023metaformer}, MS-ResNet~\citep{hu2024advancing}, SpikingResformer~\citep{shi2024spikingresformer}, and SD-Transformer v2~\citep{yao2024spike}, focusing on parameters, power consumption, and accuracy metrics. 
The comparison results are summarized in Table~\ref{table_imagenet_result}. 
Notably, with only 6.8M parameters, the QSD-Transformer achieves a top-1 accuracy of 80.3\% in the SNN domain, showcasing significant advantages in both accuracy and efficiency.
% This dataset comprises approximately 1.3M training images and 50K validation images, covering common 1000 classes.
Specifically, \textbf{QSD-Transformer} vs. SD-Transformer v2 \citep{yao2024spike} vs. SpikingResformer \citep{shi2024spikingresformer}: Param, \textbf{6.8M} vs. 55.4M vs. 60.4M; Power: \textbf{8.7mJ} vs. 52.4mJ vs. 9.7mJ; Acc, \textbf{80.3\%} vs. 79.7\% vs. 78.7\%.
When compared to the SOTA model in the SNN field, SD-Transformer v2~\citep{yao2024spike}, our method achieves a 0.6\% improvement in accuracy while reducing parameter by 87.58\% and power by 83.40\%. 
In summary, the QSD-Transformer establishes better results in both accuracy as well as efficiency on ImageNet-1K in the SNN domain and especially shines in efficiency.

%我们在准确性、参数和功效方面对提出的 QSD-Transformer 与已有的先进工作，如Metaformer、MS-ResNet、Spikformer和SpikingResformer等进行了全面比较，如表x所示。我们可以看到，QSD-Transformer 在仅用6.88M参数的情况下，在 SNN 领域实现了top-1的80.3%的准确率，这意味着我们的方法有着显著的精度和能耗优势。具体而言，\textbf{QSD-Transformer} vs. SD-Transformer v2 vs. SpikingResformer：Param，\textbf{6.88M} vs. 55.4M vs. 55.4M； Power: \textbf{8.7mJ} vs. 52.4mJ vs. 8.8mJ; Acc，\textbf{80.3\%} vs. 79.7\% vs. 78.7\%。令人惊讶的是，与SNN领域中最先进的SD-Transformer v2相比，我们的方法在参数量下降87.58%和能耗下降83.40%的前提下，还实现了0.6%的准确率提升。综上所述，QSD-Transformer 在 SNN 中的 ImageNet-1K 数据集上在精度和能耗方面都实现了SOTA，尤其在能耗方面大放异彩。

% Results on ImageNet-1K \citep{deng2009imagenet}. In all table headers, `Spike' means `Spike-driven,' and `W-A-T' represents the bit width of the weight, the bit width of the activity pattern, and the inference time step, respectively. Power refers to the average theoretical energy consumption when predicting a batch of images from the test set, with details in Appendix \ref{Energy}. $^{\star}$ indicates self-implementation results with open-source code \citep{yu2023metaformer}.
% caption: in quantization, 'W' represents the bit width of the weight, 'A' represents the bit width of the pattern, and 'T' represents the time step. Power is the average theoretical energy consumption when predicting a batch of images from the test set, details of which are shown in Eq.\ref{energy}.
% Power refers to the average theoretical energy consumption with details in Appendix \ref{Energy}.
\begin{table}[!t]
\caption{ImageNet classification results \citep{deng2009imagenet}. ` Bits' denotes the bit width of the weight and activity, respectively. Power is the estimation of energy consumption
same as \cite{yao2023spike}. $^{\star}$ indicates self-implementation results with open-source code \citep{yu2023metaformer}.}
\vspace{-2mm}
\centering
\begin{adjustbox}{max width=\linewidth} 
\begin{tabular}{cccccccc}
\toprule
    \multicolumn{1}{c}{ Method} &\multicolumn{1}{c}{  Architecture}
&\multicolumn{1}{c}{ Bits}
&\multicolumn{1}{c}{  \tabincell{c}{Spike\\-driven}}
&\multicolumn{1}{c}{ \tabincell{c}{Time\\Step}}
&\multicolumn{1}{c}{  \tabincell{c}{Param\\ (M)}}
&\multicolumn{1}{c}{  \tabincell{c}{Power\\ (mJ)}}
&\multicolumn{1}{c}{ \tabincell{c}{Acc.\\(\%)}}\\
\cline{1-8}
\multirow{1}{*}{Transformer \citep{yu2023metaformer}} 
% & {Metaformer \citep{yu2022metaformer}} & 32-32& \xmark& N/A & {21.3}   &46.5& {80.7} \\
& CAformer$^\star$ & 32-32 & \xmark& N/A & {15.1}   &40.3&79.9 \\
% \zjy{ \multirow{1}{*}{Transformer \citep{yu2023metaformer}}} &\zjy{ Q-ViT }& \zjy{ 4-4} & \zjy{ \xmark}&\zjy{  N/A }& \zjy{ {11.4} }  &\zjy{ 8.1}&\zjy{ 80.9} \\
\cline{1-8}
  QCFS \citep{bu2021optimal}  & ResNet-34 &32-1& \cmark&256 & 21.8 & - & 73.4 \\
   MST \citep{wang2023masked}   & Swin-T & 32-1& \cmark & 128 & {28.5} & {-} & {77.9} \\
   % \zjy{SpikeZIP-TF} \citep{spikeziptf2024}  & \zjy{ViT-S} & \zjy{32-1} & \cmark & \zjy{64} & \zjy{22.1} & {-} & \zjy{81.5} \\
\cline{1-8}
  \multirow{2}{*}{SEW-ResNet \citep{fang_deep_snn_2021}}
     & SEW-ResNet-34 &32-1&\xmark &4 & 25.6 &{4.9} & 67.8 \\
& SEW-ResNet-152 &32-1&\xmark &4 &60.2 &{12.9}  & 69.2 \\ \cline{1-8}

\multirow{2}{*}{MS-ResNet\citep{hu2024advancing}}
  & MS-ResNet-34 &32-1 & \cmark &4&21.8 & {5.1}  & 69.4 \\
     &MS-ResNet-104 &32-1& \cmark&4 &77.3 &{10.2}  & 75.3 \\ \cline{1-8}
\cline{1-8}
\multirow{2}{*}{Spikformer \citep{zhou2023spikformer}}
& Spikformer-8-512&32-1&\xmark&4 & 29.7 &{11.6} & 73.4\\
&Spikformer-8-768&32-1&\xmark&4 & 66.3 &{21.5}  & 74.8\\\cline{1-8}
% \multirow{1}{*}{\zjy{QKformer} \citep{zhou2024qkformer}} 
% & \zjy{HST-10-384} & \zjy{32-1} & {\cmark} & \zjy{4} & \zjy{16.4} & {-}  & \zjy{78.8} \\
\cline{1-8}

\multirow{2}{*}{SD-Transformer \citep{yao2023spike}}
& SD-Transformer-8-512&32-1&{\cmark}&4&{29.7}&{4.5}  & 74.6\\
&SD-Transformer-8-768&32-1&{\cmark}&4&{66.3}&{6.1}  & 76.3\\\cline{1-8}
\multirow{2}{*}{SpikingResformer \citep{shi2024spikingresformer}}& SpikingResformer-T &32-1&{\cmark}&{4}&{11.1} & 4.2& 74.3\\
&SpikingResformer-L&32-1&{\cmark}&4&{60.4}&{9.7}  & 78.7
\\ \cline{1-8}
\multirow{3}{*}{SD-Transformer v2 \citep{yao2024spike}}& SD-Transformer v2-T&32-1&{\cmark}&{4}&{15.1} & 16.7 & 74.1\\
&SD-Transformer v2-M&32-1&{\cmark}&4&{31.3}&{32.8}  & 77.2\\
&SD-Transformer v2-L&32-1&{\cmark}&4&{55.4}&{52.4}  & 79.7
\\ \cline{1-8}
\multirow{3}{*}{\textbf{QSD-Transformer}}& SD-Transformer v2-T&4-1&{\cmark}&{4}& 1.8 & \textbf{2.5} & 77.5\\
&SD-Transformer v2-M&4-1&{\cmark}&4&{3.9}&{5.7}  & 78.9\\
&SD-Transformer v2-L&4-1&{\cmark}&4&{6.8}&{8.7}  & \textbf{80.3}\\
% &\zjy{{HST-10-384}}&\zjy{4-1}&\zjy{{\cmark}}&\zjy{4}&\zjy{{2.3}}&\zjy{{-}}&\zjy{{\textbf{79.3}}} \\
% &\zjy{{ViT-S}}&\zjy{4-1}&\zjy{{\cmark}}&\zjy{4}&\zjy{{3.4}}&\zjy{{-}}&\zjy{{\textbf{81.9}}} \\
\bottomrule
\end{tabular}
\end{adjustbox}
\label{table_imagenet_result}
\end{table}

% \vspace{-2mm}
\begin{table*}[!t]
\centering
\caption{Object detection results on COCO 2017 \citep{lin2014microsoft}. }
%{-2mm}
\vspace{-2mm}
\begin{adjustbox}{max width=\linewidth} 
\begin{tabular}{ccccccccc}
\toprule
Method &  Architecture& Bits &\tabincell{c}{Spike\\-driven} &\tabincell{c}{Time\\Step}& \begin{tabular}[c]{@{}c@{}}Param\\      (M)\end{tabular} &  \begin{tabular}[c]{@{}c@{}}Power\\      (mJ)\end{tabular}  &  \tabincell{c}{ mAP@0.5\\(\%)} \\
\midrule
Transformer \citep{yu2023metaformer}
&CAformer & 32-32& \xmark& N/A   & 31.2 &  890.6 & 54.0   \\
Transformer \citep{zhu2020deformable}&DETR  & 32-32& \xmark& N/A   & 41.0 &  860.2 & 57.0   \\
\midrule
Spiking-Yolo \citep{kim2020spiking}
& ResNet-18 & 32-1&\cmark & 3500 & 10.2 &- & 25.7 \\ 
 Spike Calibration \citep{li2022spike} &  ResNet-18 & 32-1& \cmark  & 512 & 17.1 & -& 45.3 \\
\midrule
% % Spike Retina \citep{zhang2023direct}
% & Spike-ResNet-18& 32-1& \cmark& 4   & 11.3 & -  & 28.5     \\
EMS-SNN \citep{su2023deep}& EMS-ResNet-18& 32-1& \cmark& 4   & 26.9 & -  & 50.1     \\

 \multirow{1}{*}{\begin{tabular}[c]{@{}c@{}}SD-Transformer v2 \citep{yao2024spike}\end{tabular}}&SD-Transformer v2-M & 32-1& \cmark& 1  & 75.0 & 140.8  & 51.2    \\
\cline{1-8}
 \multirow{2}{*}{\begin{tabular}[c]{@{}c@{}} \textbf{QSD-Transformer} \end{tabular}}&SD-Transformer v2-T& 4-1 & \cmark& 4   & 16.9 & \textbf{45.1}  & 48.1     \\
&SD-Transformer v2-M& 4-1& \cmark & 4   & 34.9 &117.2   & \textbf{57.0}   \\
\bottomrule
\end{tabular}
\end{adjustbox}
\label{tab:det}
\end{table*}

\par\noindent\textbf{Object detection.}
We evaluate the efficacy of the QSD-Transformer on the object detection task and select the classic and large-scale COCO~\citep{lin2014microsoft} dataset as our benchmark for evaluation. 
% This dataset encompasses a total of 123,287 images, with 118,287 allocated for training and 5,000 for testing.
Similar to the previous work~\citep{yao2024spike}, we convert the \emph{mmdetection} \citep{chen2019mmdetection} codebase into a spiking version by IE-LIF and then use it to execute our model. We employ the QSD-Transformer as the backbone to extract features, along with Mask R-CNN \citep{he2017mask} for object detection.
The backbone is initialized with the pre-trained QSD-Transformer on ImageNet-1K, and other added layers are initialized with Xavier \citep{glorot2010understanding}.
The comparison results are summarized in Table \ref{tab:det}.
% As shown in Table \ref{tab:det}, we choose some representative and advanced deep models for comparisons, such as DETR~\citep{zhu2020deformable}, Spiking-Yolo~\citep{kim2020spiking}, and SD-Transformer v2~\citep{yao2024spike}, etc.
Obviously, the QSD-Transformer outperforms the existing state-of-the-art methods in the SNN domain by a significant margin. 
More specifically, our method exceeds the performance of SD-Transformer v2 by 5.8\% in terms of the mAP@0.5 metric, while utilizing fewer than half the parameters.
% Moreover, it is worth mentioning that although Res-SNN uses fewer parameters (26.9M) than ours, this is due to its higher training cost and special network design, and its performance lags behind that of ours by 6.9\%. 
In conclusion, our approach demonstrates efficacy in object detection tasks and has established a new benchmark for detection within the SNN domain.

\noindent\textbf{Semantic segmentation.}
We validate the efficacy of the QSD-Transformer on the semantic segmentation task and select the challenging ADE20K \citep{zhou2017scene} dataset. 
% This dataset encompasses 20K training images and 2K validation images, covering 150 categories.
Similar to the procedures in object detection, 
we converted the \emph{mmsegmentation} \citep{contributors2020mmsegmentation} codebase into a spiking version and utilized it to execute our model. The QSD-Transformer serves as the backbone for feature extraction, integrated with Semantic FPN \citep{kirillov2019panoptic} for segmentation.
The initialization is similar to that in the object detection task.
The backbone is initialized with a pre-trained model on ImageNet-1K, and the added layers are initialized using Xavier \citep{glorot2010understanding}.
Since SD-Transformer v2 is the only work in the SNN field reporting results on ADE20K, we compare our approach with advanced deep models.
As depicted in Table \ref{tab:seg}, our method significantly outperforms SD-Transformer v2 \citep{yao2024spike} across all comparison metrics, achieving an 83.94\% reduction in parameters, a 79.36\% decrease in power, and a 5.2\% increase in MIoU.
Moreover, our method achieves a comparable MIoU to the advanced DeepLab-V3 in the ANN domain while substantially reducing both parameters and power. 

\begin{table*}[!t]
\centering
\caption{Semantic segmentation results on ADE20K \citep{zhou2017scene}.}
%{-2mm}
\vspace{-2mm}
\begin{adjustbox}{max width=\linewidth} 
\begin{tabular}{cccccccc}
\toprule
Method&  Architecture  &Bits& \tabincell{c}{Spike\\-driven} &  \tabincell{c}{Time\\Step}& \begin{tabular}[c]{@{}c@{}}Param\\      (M)\end{tabular} &  \begin{tabular}[c]{@{}c@{}}Power\\      (mJ)\end{tabular}  &  \tabincell{c}{MIoU\\(\%)} \\
\midrule
Segformer \citep{xie2021segformer}
& Segformer  & 32-32 & \xmark& N/A  & 3.8 &  38.9 & 37.4   \\
DeepLab-V3 \citep{zhang2022resnest}& DeepLab-V3 & 32-32& \xmark & N/A & 68.1 & 1240.6  & 42.7   \\
\midrule
SD-Transformer v2 \citep{yao2024spike} & SD-Transformer v2-M& 32-1 & \cmark& 4  & 59.8 & 183.6  & 35.3    \\
\cline{1-8}
 \multirow{2}{*}{\begin{tabular}[c]{@{}c@{}} \textbf{QSD-Transformer} \end{tabular}}&SD-Transformer v2-T & 4-1& \cmark& 4   & 3.3 &  \textbf{17.5} & 39.0     \\
&SD-Transformer v2-M& 4-1& \cmark & 4   & 9.6 & 37.9   & \textbf{40.5}   \\
\bottomrule
\end{tabular}
\end{adjustbox}
\label{tab:seg}
\end{table*}

\vspace{-2mm}
\begin{table}[!t]
   \centering
   \caption{Transfer learning results on CIFAR10, CIFAR100 and CIFAR10-DVS.}
\vspace{-2mm}
   \begin{adjustbox}{max width=\linewidth} 
   \begin{tabular}{cccccccc}
   
   \toprule 
   
   \multirow{3}*{{ Method}}&\multirow{3}*{{\begin{tabular}[c]{@{}c@{}}Param\\ (M)\end{tabular}}} & \multicolumn{2}{c}{{CIFAR10}} & \multicolumn{2}{c}{{CIFAR100}} & \multicolumn{2}{c}{{CIFAR10-DVS}} \\
   \cmidrule(l{2pt}r{2pt}){3-4}\cmidrule(l{2pt}r{2pt}){5-6}\cmidrule(l{2pt}r{2pt}){7-8}
    &{} & {$T$} & {Acc. (\%)} & {$T$} & {Acc. (\%)} & {$T$} & {Acc. (\%)} \\
    \midrule
   %   QCFS~\citep{bu2021optimal}& -& 4 & 97.0 & 4 & 83.8 & - & -\\ 
   %  MST~\citep{wang2023masked}& - & 64 & 96.3 & 64 & 85.4 & 128 & 86.6\\ 
   % \midrule
   % Dspike~\citep{li2021differentiable} & -&6&94.3&6&75.4&10&75.4\\ 
   % Spikformer \citep{zhou2023spikformer} & - & 4 & 95.2 & 4 & 77.8 & 10 & 80.9\\ 
   % &SD-Transformer \citep{yao2023spike} & -& 4 & 95.6 & 4 &78.4 & - & -\\ 
   Spikformer \citep{zhou2023spikformer}& 29.1 & 4 & 97.0 & 4 & 83.8 & - & -\\ 
    SpikingResformer \citep{shi2024spikingresformer}&17.3 & 4 & 97.4 & 4 & 85.9 & 10 & 84.8\\ 
\midrule
    \multirow{2}{*}{\textbf{QSD-Transformer}}& 1.8 & 4 & 97.8$\pm$0.1 & 4 &86.6$\pm$0.3& 10 & 88.8$\pm$0.1 \\ 
    & 6.8 & 4 & \textbf{98.4}$\pm$0.2 & 4 & \textbf{87.6}$\pm$0.2& 10 & \textbf{89.8}$\pm$0.1 \\ 
   
   \bottomrule
   \end{tabular}
\end{adjustbox}
\label{tab:main}
\end{table}

\par\noindent\textbf{Transfer learning.}
We demonstrate the efficacy of the QSD-Transformer on transfer learning tasks. We evaluate the model's transfer learning capability on both static datasets (CIFAR) \citep{krizhevsky_2009_CIFAR10} and the neuromorphic dataset (CIFAR10-DVS) \citep{Li_Cifar10-DVS_2017} using five repeated experiments with different random seeds. To assess this ability, we fine-tune the pre-trained weights from the ImageNet-1K dataset on these selected datasets.  Compared with existing transfer learning methods in SNNs, such as SpikingResformer and Spikformer, the proposed QSD-Transformer demonstrates state-of-the-art results. It achieves 98.4\% accuracy on CIFAR-10, 87.6\% accuracy on CIFAR-100, and 89.8\% accuracy on CIFAR10-DVS, surpassing SpikingResformer by 1.0\%, 1.7\%, and 5.0\%, respectively. Thus, our method achieves the best performance across various computer vision tasks.
% We also demonstrate the efficacy of the QSD-Transformer on transfer learning tasks. We evaluate the model's transfer learning capability on both static datasets (CIFAR) and the neuromorphic dataset (CIFAR10-DVS) using five repeated experiments with different random seeds. To assess this ability, we migrate the pre-trained weights from the ImageNet-1K dataset to these selected datasets for fine-tuning. For comparison, we employ three different learning algorithms within the SNN domain: direct training methods, ANN2SNN methods, and transfer learning methods. Table~\ref{tab:main} lists the comparison results, showing that transfer learning yields relatively high accuracy among the three types of learning algorithms. 
% This superior performance can be attributed to its ability to leverage prior knowledge, facilitating adjustment to the current task. Compared with existing transfer learning methods in SNNs, such as SpikingResformer and Spikformer, the proposed QSD-Transformer demonstrates state-of-the-art results. It achieves 98.4\% accuracy on CIFAR-10, 87.6\% accuracy on CIFAR-100, and 89.8\% accuracy on CIFAR10-DVS, surpassing SpikingResformer by 1.0\%, 1.7\%, and 5.0\%, respectively. Thus, our method has demonstrated commendable performance across a variety of computer vision tasks.

% \par\noindent\textbf{IE-LIF and FGD Can Reduce Spike Information Distortion.} We e

\par\noindent\textbf{Ablation study.}
%在时间步长为4条件下进行消融
%不同权重比特数，不同模块，与ANN中QAT的量化结果对比
We first ablate two components of the QSD-Transformer, namely the IE-LIF and FGD schemes, to verify the effectiveness of the proposed method. Additionally, we quantized the weights to 4, 3, and 2 bits to study the impact of bit width on performance.
Experiments are performed on the ImageNet dataset.
The results are shown in Table \ref{tab:ablation}, where the QSD-Transformer baseline without the IE-LIF neuron and FGD scheme achieves an accuracy of 70.0\%. 
In contrast, using the IE-LIF neuron increases the accuracy by 5.8\%.
With both the IE-LIF neuron and FGD scheme, the accuracy further reaches 77.5\%.
Therefore, both the proposed IE-LIF neuron and the FGD scheme can improve performance, and their combined use can bring more significant accuracy.
Moreover, we also investigate the impact of bit-width on performance.
It can be seen from Table~\ref{tab:ablation} that the accuracy decreases with bit width reduction. 
Notably, even when the weights are quantized to 2-bit, our method still achieves 75.0\% accuracy.
\par
% Then we performed quantization on the Spikformer \citep{zhou2023spikformer} to demonstrate the robustness and scalability of our approach. The baseline refers to the Spikformer-8-384 model with a timestep of 4. Experiments are also performed
% on the ImageNet dataset. As shown in Table \ref{tab:ablation},  directly quantizing Spikformer revealed a significant performance drop of 6.14\% under standard quantization conditions. However, by applying our IE-LIF spiking neurons, we were able to improve accuracy by 6.0\%. Additionally, using both IE-LIF neurons and the FGD scheme, the accuracy further increased to 75.5\%. This shows that our method can be robustly applied to various spike-based Transformers.
Next, we delve into the application of our method within the Spikformer architecture \citep{zhou2023spikformer}  to validate its robustness and scalability. Specifically, we initially established a Spikformer-8-384 model as a quantization baseline under the conditions of a time step of 4 and a 4-bit quantization of the weights. Subsequently, we conducted ablation experiments of various modules and weight bit-widths on the ImageNet dataset. As shown in Table \ref{tab:ablation}, direct quantization of Spikformer indicates a significant performance drop of 6.14\% under standard quantization conditions. Then, by applying our IE-LIF spiking neurons, we were able to enhance the accuracy by 6.0\%. Furthermore, the accuracy was further improved to 75.5\% by combining the IE-LIF neurons with the FGD scheme. We also investigated the impact of bit-width on performance. 
Notably, our method maintains 73.1\% accuracy even with 2-bit weight quantization. The above results demonstrate that our method can be robustly applied to various spiking-based Transformer models.
\begin{table*}[!t]%{r}{0.44\textwidth}
\centering
\caption{Ablation study of the IE-LIF, FGD, and different bits.}
\vspace{-2mm}
\begin{adjustbox}{max width=\linewidth} 
\begin{tabular}{ccccc}
\toprule
Architecture    &   IE-LIF &FGD   &  Weight Bits &Acc.(\%)\\ \hline
\multirow{5}{*}{{SD-Transformer v2 \citep{yao2024spike}}}
 &- & -   & 4     & 70.0   \\
&\cmark &  -  &4&   75.8    \\
 &\cmark &\cmark     &4 &   \textbf{77.5}\\ 
&\cmark &\cmark    &3    &76.9\\
&\cmark &\cmark     &2 &  75.0   \\
\cmidrule{1-5}
\multirow{5}{*}{Spikformer \citep{zhou2023spikformer}}
 & -& -   & 4     & 64.1  \\
&\cmark & -   &4&   70.1  \\
 &\cmark &\cmark     &4 &   \textbf{75.5}\\ 
&\cmark &\cmark    &3    &74.1\\
&\cmark &\cmark     &2 &  73.1   \\
    \bottomrule
\end{tabular}
\end{adjustbox}
\label{tab:ablation}
\end{table*}
\par
\begin{wraptable}[12]{r}{0.45\textwidth} 
\vspace{-0.3cm}
\caption{Ablation study of the activity bits and training time step on the QSD-Transformer.}
\centering
\begin{adjustbox}{max width=\linewidth} 
\begin{tabular}{ccc}
\toprule
\multicolumn{1}{c}{Bits ($b$)}
&\multicolumn{1}{c}{Timestep ($T$)}
&\multicolumn{1}{c}{{Acc.(\%)}}\\
\cline{1-3}
  1 & 1   & {67.6} \\
 1 & 2   & {68.5} \\
 1 & 4   & {70.0} \\
 2 & 1   & {71.6} \\
 2 & 2   & {77.4} \\
 4 & 1   & \textbf{77.5} \\
\bottomrule
\end{tabular}
\end{adjustbox}
\label{bits}
\end{wraptable}
% Finally, we conducted ablation experiments on the activity bit $b$ and training time step $T$ of IE-LIF.  
% As shown in  Table \ref{bits}, increasing the IE-LIF activity bit significantly improves performance. With $T=1$, increasing $b$ from 1 to 4 improves accuracy by 9.9\%; with $b=1$, increasing from 1 to 4 only improves accuracy by 2.4\%. This is because increasing activity bits $b$ enhances the information capacity and reduces quantization errors while increasing training time step $T$ has a smaller effect due to numerous redundancy spike trains. These results show that quantization performance has a greater impact on the activity bits than the setting of time steps.
Finally, our ablation studies on the activity bit \( b \) and training time step \( T \) of the IE-LIF model reveal that augmenting the activity bit \( b \) substantially boosts performance. As depicted in Table \ref{bits}, elevating \( b \) from 1 to 4 with \( T=1 \) results in a 9.9\% increase in accuracy; conversely, with \( b=1 \), raising \( T \) from 1 to 4 yields a more modest 2.4\% improvement. This disparity arises because augmenting the activity bits \( b \) enhances the model's information capacity and mitigates quantization errors, whereas increasing the training time step \( T \) has a less pronounced impact, likely due to the redundancy inherent in spike trains. Furthermore, extending the time step \( T \) incurs significant memory and energy costs, which is not the case for increasing the activity bit \( b \). These findings underscore that the quantization performance is more sensitive to the activity bits than the time step configuration.
\vspace{-3mm}
\section{Conclusion}
% Recently, researchers in the SNN domain have been dedicated to designing large and complex Spiking Transformer architectures.
% Despite achieving high performance, these large-scale structures come with massive parameters and computational resources, severely limiting their deployment on resource-constrained edge platforms.
In this paper, we first introduce the lightweight spike-driven transformer, namely the QSD-Transformer, which quantifies the weights from 32-bit to low-bit. By employing both low-bit weights and 1-bit spike activities, QSD-Transformer has demonstrated significant energy efficiency.
% Unfortunately, the bimodal distribution of the information in Q-SDSA is highly sensitive to the quantization operation, leading to the performance degradation of the QSD-Transformer.
Despite exhibiting efficiency benefits, the QSD-Transformer suffers from performance degradation.
We reveal that this is attributed to the SID problem and propose a bi-level optimization strategy to solve this challenge.
At the lower level, we propose the IE-LIF neuron, which generates multi-bit spikes in training while maintaining spike-driven behavior during inference.
At the upper level, we introduce the FGD scheme, which optimizes attention distribution between the Q-SDSA and its ANN counterpart.
Extensive experiments show that our method achieves state-of-the-art results in both performance and efficiency on various vision tasks, paving the way for the practical deployment of spike-based Transformers in resource-limited platforms.

\section*{Acknowledgement}
This work was supported in part by the National Natural Science Foundation of China under grant U20B2063, 62220106008, and 62106038, the Sichuan Science and Technology Program under Grant 2024NSFTD0034 and 2023YFG0259. The research is supported by Shenzhen Science and Technology Program ZDSYS20230626091302006; and Shenzhen Science and Technology Research Fund (Fundamental Research Key Project Grant No. JCYJ20220818103001002.
\bibliography{iclr2025_conference}
\bibliographystyle{iclr2025_conference}
%{50mm}
\newpage
\appendix
% \vspace{2mm}
\textbf{\large{Appendix}}

\section{Backpropagation process of spiking neurons}
\label{bp}
There exist two primary methods of training high-performance SNNs. One way is to discretize ANN into spike form through neuron equivalence \citep{li2021free,bu_2022_optimized,hao2023reducing,ding2021optimal}, i.e., ANN-to-SNN conversion, but this requires a long simulation time step and boosts the energy consumption. We employ the direct training method \citep{wu2018spatio,Qiu2024,wei2024event} and apply surrogate gradient training. 
 \par
Then in this section, we introduce the training process of SNN gradient descent and the parameter update method of spatio-temporal backpropagation (STBP) \citep{wu2018spatio,xiao2022online}. SNNs' parameters can be taught using gradient descent techniques, just like ANNs, after determining the derivative of the generation process. Moreover, the accumulated gradients of loss $\mathcal{L}$ with respect to weights $\mathbf{w}$ at layer $\ell$ can be calculated as:
\begin{equation}
\small
    \frac{\partial \mathcal{L}}{\partial \mathbf{w}^{\ell}}=\sum_{t=1}^T \frac{\partial \mathcal{L}}{\partial \mathbf{s}^{\ell+1}[t]}\frac{\partial \mathbf{s}^{\ell+1}[t]}{\partial \mathbf{v}^{\ell+1}[t]}\left(\frac{\partial \mathbf{v}^{\ell+1}[t]}{\partial \mathbf{w}^\ell}+\sum_{\tau < t}\prod_{i=t-1}^{\tau}\left({\frac{\partial \mathbf{v}^{\ell+1}[i+1]}{\partial \mathbf{v}^{\ell+1}[i]}}+{\frac{\partial \mathbf{v}^{\ell+1}[i+1]}{\partial \mathbf{s}^{\ell+1}[i]}}{\frac{\partial \mathbf{s}^{\ell+1}[i]}{\partial \mathbf{v}^{\ell+1}[i]}}\right)\frac{\partial \mathbf{v}^{\ell+1}[\tau]}{\partial \mathbf{w}^\ell}\right),
    \label{bptt gradient}
\end{equation}
where $\mathbf{s}^{\ell}[t]$ and $\mathbf{v}^{\ell}[t]$ represent the binary and membrane potential of the neuron in layer $\ell$, at time $t$. Moreover, notice that $ \frac{\partial \mathbf{s}^{\ell}[t]}{\partial \mathbf{v}^{\ell}[t]}$ is non-differentiable. To overcome this problem, \citep{wu2018spatio} propose the surrogate function to make only the neurons whose membrane potentials close to the firing threshold receive nonzero gradients during backpropagation.  In this paper, we use the rectangle function, which has been shown to be effective in gradient descent and may be calculated by:
\begin{equation}
\label{eq3}
    \frac{\partial \mathbf{s}^{\ell}[t]}{\partial \mathbf{v}^{\ell}[t]}=\frac{1}{a} \operatorname{sign}\left(\left|\mathbf{v}^{\ell}[t]-\vartheta \right|<\frac{a}{2}\right),
\end{equation}
where $a$ is a defined coefficient for controlling the width of the gradient window.

\section{Proof of the Proposition 1.}
\label{lim}
\par
\noindent \textbf{Proposition 1.} \textit{
Given the SNN Transformer and ANN Transformer models, where the distributions of the query ($\mathbf{q}$), key ($\mathbf{k}$), and value ($\mathbf{v}$) follow binomial $\mathcal{B}(r, T)$ and normal $\mathcal{N}(\mu,\sigma)$ distributions, respectively, it is postulated that as the SNN's time step $T$ tends to infinity, there exist parameters $\mu$, $\sigma$, and $r$ such that the average entropy over time of the SNN's attention scores $\mathcal{H}(\sum_{t=1}^{T}\mathbf{p}^\mathcal{S}[t])$ equals ANN attention scores' entropy $\mathcal{H}(\mathbf{p}^\mathcal{A})$.}
\begin{proof}
\textbf{Proposition 1.} can be restated as follows:
 \begin{equation}
     \lim_{T \to \infty} \exists \mu,\sigma, r \quad \mathcal{H}(\sum_{t=1}^{T}\mathbf{p}^\mathcal{S}[t])=\mathcal{H}(\mathbf{p}^\mathcal{A}),
\end{equation}   
where ${\bf p}^{\mathcal{A}}$ and ${\bf p}^{\mathcal{S}}$ represent the query $\mathbf{q}$, key $\mathbf{k}$, and value $\mathbf{v}$ in the same architecture ANN (teacher) and QSD-Transformer (student), and following the binomial $\mathcal{B}(r, T)$ and normal $\mathcal{N}(\mu,\sigma)$ distributions. $\theta^{\mathcal{S}}$ is the parameters of the student (QSD-Transformer). 
\par
Assume $\mathbf{p}^\mathcal{S}[1], \mathbf{p}^\mathcal{S}[2], \mathbf{p}^\mathcal{S}[3], \ldots, \mathbf{p}^\mathcal{S}[t]$ are $t$ independent random variables, each following a binomial distribution. The expectation is $\mathbb{E}(\mathbf{p}^\mathcal{S}[t]) = r^\mathcal{S}[t]$, where $r^\mathcal{S}[t]$ is the firing rate of the SNN at time $t$. The variance is given by $\mathbb{D}(\mathbf{p}^\mathcal{S}[t]) = \sigma^\mathcal{S}[t]$. And
Let $ \mathbf{y}^\mathcal{S}[t] = \mathbf{p}^\mathcal{S}[t] - r^\mathcal{S}[t]$, where $ \mathbb{E}(\mathbf{y}^\mathcal{S}[t]) = 0$ and $\mathbb{D}(\mathbf{y}^\mathcal{S}[t]) = \sigma$. Let the characteristic function \citep{chow2012probability} of the random variable $ \mathbf{y}^\mathcal{S}[t] $ be $ \varphi_{\mathbf{y}^\mathcal{S}[t]}(j) $.
Then let the random variable $ \mathbb{\eta} = \frac{\mathbf{y}^\mathcal{S}[1]+\mathbf{y}^\mathcal{S}[2]+ \mathbf{y}^\mathcal{S}[3]+\ldots+\mathbf{y}^\mathcal{S}[T]}{\sqrt{t\sigma}} $. Then the characteristic function of $ \eta $ is:
\begin{equation}
    \varphi_{\eta}=\left[\varphi_{\mathbf{y}^\mathcal{S}[t]}(\frac{j}{\sqrt{T\sigma}})\right]\cdot\left[\varphi_{\mathbf{y}^\mathcal{S}[t]}(\frac{j}{\sqrt{T\sigma}})\right]\ldots\left[\varphi_{\mathbf{y}^\mathcal{S}[t]}(\frac{j}{\sqrt{T\sigma}})\right]=\left[\varphi_{\mathbf{y}^\mathcal{S}[t]}(\frac{j}{\sqrt{T\sigma}})\right]^{T},
\end{equation}
Then when SNN's timestep $ T \to \infty $, $\frac{j}{\sqrt{T\sigma}}$ can be expanded at the point 0 using the Taylor series:

\[ \varphi_{\mathbf{y}^\mathcal{S}[t]}(\frac{j}{\sqrt{T\sigma}})) = \varphi_{\mathbf{y}^\mathcal{S}[t]} (0) + \varphi_{\mathbf{y}^\mathcal{S}[t]}' (0) \left( \frac{j}{\sqrt{T\sigma}} \right) + \frac{\varphi_{\mathbf{y}^\mathcal{S}[t]}'' (0)}{2!} \left(\frac{j}{\sqrt{T\sigma}} \right)^2 + o\left( \left( \frac{j}{\sqrt{T\sigma}} \right)^2 \right), \]

Since $\varphi_{\mathbf{y}^\mathcal{S}[t]} (0) = 1$, $\varphi_{\mathbf{y}^\mathcal{S}[t]}' (0)  = 0$, and $\varphi_{\mathbf{y}^\mathcal{S}[t]}'' (0) = -\sigma$, we have:

\[ \varphi_{\mathbf{y}^\mathcal{S}[t]}(\frac{j}{\sqrt{T\sigma}}) = 1 - \frac{j^2}{2T} + o\left( \left( \frac{j}{\sqrt{T\sigma}} \right)^2 \right), \]

\[ \left[\varphi_{\mathbf{y}^\mathcal{S}[t]}(\frac{j}{\sqrt{T\sigma}})\right]^{T}= \left[ 1 - \frac{j^2}{2T} + o\left( \left( \frac{j}{\sqrt{T\sigma}} \right)^2 \right) \right]^T = \left[ 1 - \frac{j^2}{2T} + o\left( \left( \frac{j}{\sqrt{T\sigma}} \right)^2 \right) \right], \]
Hence:
\begin{equation}
    \lim_{T \to \infty}\left[\varphi_{\mathbf{y}^\mathcal{S}[t]}(\frac{j}{\sqrt{T\sigma}})\right]^{T}=\lim_{T \to \infty}\left[ 1 - \frac{j^2}{2T} + o(( \frac{j}{\sqrt{T\sigma}})\right]^{T}=e^{-\frac{j^2}{2}},
\end{equation}
where $ e^{-\frac{j^2}{2}} $ happens to be the characteristic function of a random variable following the standard normal distribution $ \mathcal{N}(0, 1) $, so $\eta$ follows the standard normal distribution, which distribution is the same to the attention score in ANN Transformer. Hence, as the SNN's time step $T$ tends to infinity, $\exists \mu,\sigma, r $ such that $\mathcal{H}(\sum_{t=1}^{T}\mathbf{p}^\mathcal{S}[t])=\mathcal{H}(\mathbf{p}^\mathcal{A})$.
\end{proof}
%假设 \mathbf{p}^\mathcal{S}[1],\mathbf{p}^\mathcal{S}[2],\mathbf{p}^\mathcal{S}[3],\codts \mathbf{p}^\mathcal{S}[t]为t个独立且都服从二项分布的随机变量，其中期望E(\mathbf{p}^\mathcal{S}[t])=r^\mathcal{S}[t],其中r为t时刻SNN的发放率，方差为D(\mathbf{p}^\mathcal{S}[t])=\sigma^\mathcal{S}[t]
%不妨设 Y^\mathcal{S}[t]=\mathbf{p}^\mathcal{S}[t]-r^\mathcal{S}[t],其中E(Y^\mathcal{S}[t])=r^\mathcal{S}[t]，D(Y^\mathcal{S}[t])=\sigma^\mathcal{S}[t]
% 之后我们计算变量\mathbf{y}^\mathcal{S}[t]的特征函数
%二项分布在实验次数趋近与无穷时，等价于正态分布

\section{Proof of the Proposition 2.}
\label{entropy}
\par\noindent \textbf{Proposition 2.} \textit{For a random variable $\mathbf{x} \sim \mathcal{N}(\mu,\sigma)$, the information entropy $\mathbf{x}$ reaches its maximum value $ \mathcal{H}(\mathbf{x}) =\frac{1}{2}\log 2\pi e \sigma^{2}(\mathbf{x}) $ and is observed to increase with the expansion of variance $\sigma$.}
\par
\begin{proof}
For a continuous random variable $\mathbf{x}$ obeying a normal distribution, its probability density function $p(x)$ is given by:%, the differential entropy of $X$ can be calculated as
\begin{equation}p(x)=\frac{1}{\left(2\pi\sigma^{2}\right)^{1/2}}\exp\left\{-\frac{\left(x-\mu\right)^{2}}{2\sigma^{2}}\right\},\end{equation}
Consequently, the differential entropy of $\mathbf{x}$ can be calculated as
\begin{equation}\begin{aligned}
\mathcal{H}(\mathbf{x})&=-\int_{-\infty}^{\infty}p(x)\log p(x)dx,\\
&=-\int\frac{1}{(2\pi\sigma^{2})^{1/2}}\exp\left\{-\frac{(x-\mu)^{2}}{2\sigma^{2}}\right\}\log\frac{1}{(2\pi\sigma^{2})^{1/2}}\exp\left\{-\frac{(x-\mu)^{2}}{2\sigma^{2}}\right\}dx, \\
&=-\frac{1}{\left(2\pi\sigma^{2}\right)^{1/2}}\int\exp\left\{-\frac{\left(x-\mu\right)^{2}}{2\sigma^{2}}\right\}\left(-\log\left(\sqrt{2\pi}\sigma\right)-\frac{\left(x-\mu\right)^{2}}{2\sigma^{2}}\right)dx, \\
&=-\frac{1}{(2\pi\sigma^{2})^{1/2}}\cdot-\log\left(\sqrt{2\pi}\sigma\right)\int\exp\left\{-\frac{(x-\mu)^{2}}{2\sigma^{2}}\right\}dx+\\
&\quad\quad\frac{1}{(2\pi\sigma^{2})^{1/2}}\int\exp\left\{-\frac{(x-\mu)^{2}}{2\sigma^{2}}\right\}\frac{(x-\mu)^{2}}{2\sigma^{2}}dx, \\
&=\frac{\log\left(\sqrt{2\pi}\sigma\right)}{\left(2\pi\sigma^{2}\right)^{1/2}}\int\exp\left\{-\frac{\left(x-\mu\right)^{2}}{2\sigma^{2}}\right\}dx+\frac{1}{\left(2\pi\sigma^{2}\right)^{1/2}}\int\exp\left\{-\frac{\left(x-\mu\right)^{2}}{2\sigma^{2}}\right\}\frac{\left(x-\mu\right)^{2}}{2\sigma^{2}}dx,\\
&=\frac{\log\left(\sqrt{2\pi}\sigma\right)}{(2\pi\sigma^{2})^{1/2}}\sqrt{2}\sigma\int\exp\left\{-\left(\frac{x-\mu}{\sqrt{2}\sigma}\right)^{2}\right\}d\left(\frac{x-\mu}{\sqrt{2}\sigma}\right)+\\
&\quad\quad\frac{1}{(2\pi\sigma^{2})^{1/2}}\sqrt{2}\sigma\int\exp\left\{-\left(\frac{x-\mu}{\sqrt{2}\sigma}\right)^{2}\right\}\frac{(x-\mu)^{2}}{2\sigma^{2}}d\left(\frac{x-\mu}{\sqrt{2}\sigma}\right) .\\
\end{aligned}
\end{equation}
Moreover, it can be easily proven that
\begin{equation}\int_{-\infty}^{\infty}e^{-y^{2}}dy=\sqrt{\pi}.\end{equation}
Thus,
\begin{equation}\begin{aligned}
\mathcal{H}(\mathbf{x})
&=\frac{\log\left(\sqrt{2\pi}\sigma\right)}{\sqrt{\pi}}\int_{-\infty}^{\infty}e^{-y^{2}}dy+\frac{1}{\sqrt{\pi}}\int_{-\infty}^{\infty}e^{-y^{2}}y^{2}dy, \\
&=\log\left(\sqrt{2\pi}\sigma\right)+\frac{1}{\sqrt{\pi}}\cdot-\frac{1}{2}\biggl(0-\int_{-\infty}^{\infty}e^{-y^{2}}dy\biggr), \\
&=\log\left(\sqrt{2\pi}\sigma\right)+\frac12 , \\
&=\frac{1}{2}\Big(\log\Big(2\pi\sigma^{2}\Big)+1\Big), \\
&=\frac{1}{2}\log\Big(2\pi e\sigma^{2}\Big). \\
\end{aligned}
\end{equation}
\end{proof}

\section{Theoretical analysis in fusion of MPRF and weights}
\label{Fusion}
% In Section \ref{iesn}, we propose a membrane potential rectify function (MPRF) $\phi^{\ell}(\cdot)$ to maximize the information entropy of attention score. The homogeneity of convolution allows the following BN and linear scaling transformation to be equivalently fused into the convolutional layer with an added bias when deployment. By employing this strategy, The model performs inference in a very short period without additional computational overhead. To be specific, we use Eq. \eqref{qconv_formula} to express the quantized convolution (Q-Conv):

In Section \ref{iesn}, we introduce a membrane potential rectify function (MPRF) $\phi^{\ell}(\cdot)$ aimed at maximizing the information entropy of the attention score. The inherent homogeneity of convolution operations permits the subsequent batch normalization (BN) and linear scaling transformations to be seamlessly integrated into the convolutional layer with an added bias during deployment. This approach enables the model to conduct inference rapidly without incurring additional computational overhead. Specifically, we utilize Eq. \ref{qconv_formula} to represent the quantized convolution (Q-Conv):

\begin{equation}
    \label{qconv_formula}
    \mathbf{y}_Q^{\ell} = \mathbf{w}_{Q_{Conv}}^{\ell} \cdot \mathbf{S}^{\ell} + \mathbf{b}_{Q_{Conv}}^{\ell}
\end{equation}

where $\mathbf{S}$ denotes input binary spike, $\mathbf{w}_{Q_{Conv}}$ and $\mathbf{b}_{Q_{Conv}}$ are quantized weights and bias of the Q-Conv layer. $\mathbf{y}_Q$ denotes the output of the Q-Conv layer. After employing MPRF, the rectified output should be computed as Eq. \ref{fusion_func}

\begin{equation}
\label{fusion_func}
    \begin{aligned}
        \hat{\mathbf{y}}_Q^{\ell} &= \phi(\mathbf{y}_Q^{\ell}) = \phi(\mathbf{w}_{Q_{Conv}}^{\ell} \cdot \mathbf{S}^{\ell} + \mathbf{b}_{Q_{Conv}}^{\ell}), \\
        &= \frac{(\mathbf{w}_{Q_{Conv}}^{\ell} \cdot \mathbf{S}^{\ell} + \mathbf{b}_{Q_{Conv}}^{\ell})-\mu(\mathbf{w}_{Q_{Conv}}^{\ell} \cdot \mathbf{S}^{\ell} + \mathbf{b}_{Q_{Conv}}^{\ell})}{\sigma(\mathbf{w}_{Q_{Conv}}^{\ell} \cdot \mathbf{S}^{\ell} + \mathbf{b}_{Q_{Conv}}^{\ell})}\cdot \mathbf{\gamma}^{\ell}  +\mathbf{\alpha}^{\ell}, \\
        &= \frac{\mathbf{\gamma}^{\ell} \cdot (\mathbf{w^{\ell}_{Q_{Conv}}} \cdot \mathbf{S}^{\ell} + \mathbf{b}^{\ell}_{Q_{Conv}})}{\mathbf{\sigma}(\mathbf{w}_{Q_{Conv}}^{\ell} \cdot \mathbf{S}^{\ell} + \mathbf{b}_{Q_{Conv}}^{\ell})} - \frac{\mathbf{\gamma}^{\ell}\cdot\mu(\mathbf{w}_{Q_{Conv}}^{\ell} \cdot \mathbf{S}^{\ell} + \mathbf{b}_{Q_{Conv}}^{\ell})}{\mathbf{\sigma}(\mathbf{w}_{Q_{Conv}}^{\ell} \cdot \mathbf{S}^{\ell} + \mathbf{b}_{Q_{Conv}}^{\ell})} + \mathbf{\alpha^{\ell}}, \\
        &= \frac{\mathbf{\gamma}^{\ell} \cdot \mathbf{w}_{Q_{Conv}}^{\ell}}{\mathbf{\sigma}(\mathbf{w}_{Q_{Conv}}^{\ell} \cdot \mathbf{S}^{\ell} + \mathbf{b}_{Q_{Conv}}^{\ell})} \cdot \mathbf{S}^{\ell} + [\frac{\mathbf{\gamma}^{\ell} \cdot \mathbf{b}_{Q_{Conv}}^{\ell} - \mu(\mathbf{w}_{Q_{Conv}}^{\ell} \cdot \mathbf{S}^{\ell} + \mathbf{b}_{Q_{Conv}}^{\ell})}{\mathbf{\sigma}(\mathbf{w}_{Q_{Conv}}^{\ell} \cdot \mathbf{S}^{\ell} + \mathbf{b}_{Q_{Conv}}^{\ell})} + \mathbf{\alpha}^{\ell}], \\
        &= \mathbf{w_{f}^{\ell}} \cdot \mathbf{S} + \mathbf{b_{f}^{\ell}},
    \end{aligned}
\end{equation}
where $\mathbf{w_f^{\ell}}$ and $\mathbf{b_f^{\ell}}$ denote the fusioned weight and bias:
\begin{equation}
\begin{aligned}
    &\mathbf{w_{f}^{\ell}} = \frac{\mathbf{\gamma}^{\ell} \cdot \mathbf{w}_{Q_{Conv}}^{\ell}}{\mathbf{\sigma}(\mathbf{w}_{Q_{Conv}}^{\ell} \cdot \mathbf{S}^{\ell} + \mathbf{b}_{Q_{Conv}}^{\ell})}, \\
    &\mathbf{b_{f}^{\ell}} = \frac{\mathbf{\gamma}^{\ell} \cdot \mathbf{b}_{Q_{Conv}}^{\ell} - \mu(\mathbf{w}_{Q_{Conv}}^{\ell} \cdot \mathbf{S}^{\ell} + \mathbf{b}_{Q_{Conv}}^{\ell})}{\mathbf{\sigma}(\mathbf{w}_{Q_{Conv}}^{\ell} \cdot \mathbf{S}^{\ell} + \mathbf{b}_{Q_{Conv}}^{\ell})} + \mathbf{\alpha}^{\ell}.
\end{aligned}
\end{equation}

% \begin{equation}
% \begin{aligned}
%     E_{tot} &= \sum_{i} E_{{AC}_i} \cdot T_i \cdot fr_i \cdot SOP_{{Conv}_i} \\
%     &= \sum_{j} E_{{AC}_j} \cdot (SOP_{q,k,v} + SOP_{f(q,k,v)} + SOP_{Linear})_j
% \end{aligned}
% \label{etot}
% \end{equation}

% where $E_{{AC}_i} = E_{{AC}_j} = \frac{1}{4}E_{{AC}_{fp32}} = 0.225pJ$, and the value of $i$ and $j$ are related to the Conv-based
% SNN block and Transformer-based
% SNN block numbers respectively.

\section{Theoretical  energy consumption analysis}
\label{Energy}
% 在不考虑实际芯片/硬件平台的制作工艺、数据访问、存储等方面的能耗影响因素，对比不同模型的计算能耗是有说服力的。这种对比有效反映了不同网络模型在其本身的计算过程中的综合效率。前人工作（Mark Horowitz. 1.1 computing’s energy problem (and what we can do about it). In 2014 IEEE International Solid-State Circuits Conference Digest of Technical Papers (ISSCC), pages 10–14. IEEE, 2014.）指出，45nm工艺的硬件平台中一次32位浮点加法的功耗是0.9pJ，一次MAC的功耗是4.6pJ(EM = 3.7pJ and EAC = 0.9pJ)。诸多对于SNN网络研究工作中的性能分析也是参考的此数据。
When disregarding the energy consumption factors related to hardware manufacturing processes, data access, and storage, comparing the computational energy consumption of different models remains compelling. Such comparisons effectively reflect the intrinsic computational efficiency of various network models. Previous work by \citep{6757323}  indicates that, on a 45nm process hardware platform, the energy consumption for a single multiply-accumulate (MAC) operation is 4.6pJ (with 3.7pJ for multiplication and 0.9pJ for addition). Many performance analyses in the research of spiking neural networks (SNNs) \citep{panda_2020_toward,qiu2024efficient,shan2024advancing} also reference this data.

\subsection{Comparision on MHSA and SDSA}

Given a float-point input sequence $X\in\mathbb{R}^{N\times D}$, the float-point Query ($\mathbf{q}$), Key ($\mathbf{k}$), and Value ($\mathbf{v}$) in $\mathbb{R}^{N\times D}$ are computed using three learnable linear matrices, where $N$ is the token number, and $D$ is the channel dimension. The MHSA scaled dot-product is computed as described by \citep{dosovitskiy2020image}:
\begin{equation}
\mathrm{MHSA}(\mathbf{q},\mathbf{k},\mathbf{v})=\mathbf{softmax}\left(\frac{\mathbf{q}\mathbf{k}^\mathrm{T}}{\sqrt{d}}\right)\mathbf{v}
\end{equation}

where $d=\frac{D}{H}$ is the feature dimension of one head and H is the number of heads, and $\sqrt{d}$ serves as the scaling factor. Typically, MHSA divides $\mathbf{q}$, $\mathbf{k}$ and $\mathbf{v}$ into $H$ heads along the channel dimension. For the $i^{th}$ head, $\mathbf{q_i}$, $\mathbf{k_i}$ and $\mathbf{v_i}$ are in $\mathbb{R}^{N\times D/H}$. After performing the self-attention operation on each of the $H$ heads independently, the outputs are concatenated together.

In MHSA, $\mathbf{q}$ and $\mathbf{k}$ are matrix-multiplied, followed by a matrix multiplication of their output with $\mathbf{v}$. The computational complexity of MHSA(·) is $O(N^2D)$, indicating a \textit{quadratic} relationship with the token number $N$. For the SDSA modules, the computational complexity is $O(ND^{2})$, and the energy cost of the Rep-Conv part is consistent with SNN-based convolution. The energy cost of the SDSA operator part is given in Table \ref{sdsacost}.

% 对比MHSA和SDSA：

\begin{table}[h]
\caption{Theoretical FLOPs/SOPs of self-attention modules.}
    \centering
    \begin{tabular}{c|c|c}
        \hline
                & Multi-head Self-attention (MHSA) & Spike-driven Self-attention (SDSA)    \\
        \hline
        Function   & $\mathrm{MHSA}(\mathbf{q},\mathbf{k},\mathbf{v})=\mathbf{softmax}\left(\frac{\mathbf{q}\mathbf{k}^\mathrm{T}}{\sqrt{d}}\right)\mathbf{v} $     &             $
                    \mathrm{SDSA}(\mathbf{q_s},\mathbf{k_s},\mathbf{k_s})=\mathcal{SN}_s((\mathbf{q_s}\mathbf{k_s}^\mathrm{T})\mathbf{v_s})
                $ \\
        \hline
     $q,k,v$    & $3ND^2$      & $T \cdot fr_1 \cdot 3 \cdot FL_{Conv}$           \\
     $f(q,k,v)$ & $2N^2D$      &  $T \cdot fr_2 \cdot ND^2$          \\
     Scale      & $N^2$      &      -         \\
     Softmax    & $2N^2$      &      -     \\
     Linear     & $FL_{fc}$      &   $T \cdot fr_3 \cdot FL_{fc}$         \\
        \hline

    \end{tabular}
\label{sdsacost}
\end{table}

\subsection{Theoretical energy consumption of QSD-Transformer}
We first  calculate the theoretical energy consumption requires calculating the synaptic operations (SOPs):
\begin{equation}
\label{etot}
\text{SOPs}^{\ell}=fr^{\ell} \times T \times \text{FLOPs}^{\ell}
\end{equation}
where $fr^\ell$, $ \text{FLOPs}^\ell$, and $T$ is the firing rate, float-pointing operations, and timestep of layer $\ell$. Moreover, the respective number of FLOPs adds $\{\frac{1}{32},\frac{1}{16},\frac{1}{8}\}$ of the number of $\{$2,3,4$\}$-bit multiplications equals the OPs following \citep{xu2023q,QIN2020107281}. 

% % 由于本文的工作中，网络的参数被量化到了4bit的精度，参与加法运算的数的位宽降低，单次加法的能耗应当相应降低。在\citep{xu2023q}工作中，估算的时候以fp32精度为基准，相加一个32bit的数和相加四次8bit的数在计算能耗上是等价的。因此我们依然可以使用fp32的能耗指标。对于整体的网络，其能耗计算公式为：
% Due to the quantization of network parameters to 4-bit precision in this work, the bit-width of the numbers involved in addition operations is reduced, and consequently, the energy consumption per addition should also decrease. In \citep{xu2023q}, it is estimated that the energy consumption of adding one 32-bit number is equivalent to adding four 8-bit numbers. Therefore, we can still use the energy consumption metrics of 32-bit floating-point (fp32) operations. 
The total energy consumption of the network can be calculated using Eq. \ref{energy} for non-quantized models and Eq. \ref{energy_qsdsa} for quantized models:
\begin{equation}
\label{energy}
    \begin{aligned}
E_{total}=E_{MAC}\cdot \text{FLOPs}_{conv}^1 + E_{AC}\cdot T \cdot(\sum_{n=2}^N \text{FLOPs}_{conv}^n \cdot fr^{n} +\sum_{m=1}^M \text{FLOPs}_{fc}^m \cdot fr^{m}),
\end{aligned}
\end{equation}

\begin{equation}
\label{energy_qsdsa}
    \begin{aligned}
E_{total}=E_{MAC}\cdot \text{FLOPs}_{conv}^1 + E_{AC}\cdot (\sum_{n=2}^{N} \text{SOPs}^{N} + \sum_{m=1}^{M} \text{SOPs}^{M})
\end{aligned}
\end{equation}

\par
where $N$ and $M$ are the total number of Conv and FC layers, $E_{MAC}$ and $E_{AC}$ are the energy costs of MAC and AC operations, and $fr^{m}$, $fr^{n}$, $\text{FLOPs}_{conv}^n$ and $\text{FLOPs}_{fc}^m$ are the firing rate and FLOPs of the $n$-th Conv and $m$-th FC layer.  Previous SNN works  \citep{6757323, rathi2021diet, hu2024high} assume 32-bit floating-point implementation in 45nm technology, where $E_{MAC}$ = 4.6pJ and $E_{AC}$ = 0.9pJ for various operations. 
\section{Limitations and Future Works}
\label{Limitations}
\paragraph{Limitations} The limitations of this work include the scalability of low-bit spike-driven Transformer models and the hardware deployment, which we will address in future research. %硬件没做的问题
The experimental results presented in this paper are reproducible. Detailed explanations of model training and configuration are provided in the main text and supplemented in the appendix. Our codes and models will be made available on GitHub after review.

\paragraph{Future Works} Since the largest Spikformer-V2 \cite{zhou2024spikformer} model has not yet released its training code and weights, we will attempt to quantify the Spikformer-V2 model in the future to demonstrate the scalability of our approach. Moreover, we did not take the energy consumption of memory access into account when calculating the theoretical energy consumption owing to the diversity of different dataflow and memory access schemes and the implementation on various hardware platforms. We will deploy our lightweight model into hardware platforms such as Field Programmable Gate Arrays (FPGAs) to evaluate the factual performance, where we will optimize the suitable read-write data streams and memory access schemes to enhance the inference speed of the models. 

\section{Experiment Details}\label{details}
%鸿霖
\subsection{ImageNet-1K experiments}\label{appendix_training_image}
%对于imagenet数据集上的实验，我们对训练数据使用了数据增强，与cite{yao2024spike}一致，包括标准预处理、归一化和随机裁剪。标准预处理包括随机增强和mixup、cutmix以及label smoothing。在训练过程中，我们设置batchsize为128。我们使用AdamW优化器，并设置weight decay为0.01。在训练的320个epoch中，学习率使用余弦衰减并被初始化为0.0001。
%For the experiments on the ImageNet dataset,我们使用如表\ref{table_train_imagenet_detail}所示的超参数.并且，我们在三种不同大小的模型上验证了我们的结果，具体模型结构如表\ref{table_imagenet_config}所示
ImageNet-1K dataset is commonly used for computer vision tasks. It spans 1000 object classes and contains around 1.3 million training images and 50,000 validation images. For experiments on the ImageNet dataset, we used the hyper-parameters shown in Table \ref{table_train_imagenet_detail}. Moreover, we employ our model on three different scales, with the specific model configurations detailed in Table \ref{table_imagenet_config}. We conducted training on eight 40GB A100 GPUs. For the three different model scales—1.8M, 3.8M and 6.8M parameters—we allocated 24, 28 and 36 hours of training time, respectively.

\begin{table}[h]
%{-0.2cm}
\centering
\caption{Hyper-parameters for image classification on ImageNet-1K and CIFAR10/100.}
\label{table_train_imagenet_detail}
% %{3pt}
\begin{tabular}{c|c|c}
\hline
Hyper-parameter     & ImageNet  & CIFAR10/10    \\ \hline
Timestep (Training/Inference)          & 1/4  & 1/4                      \\
Epochs              & 300      & 100                     \\
Resolution          & 224$\times$224   &128$\times$128        \\
Batch size          & 1568          &256             \\
Optimizer           & LAMB           & LAMB          \\
Base Learning rate  & 6e-4         & 6e-4      \\
Learning rate decay & Cosine     & Cosine \\
Warmup eopchs       & 10      & None                     \\
Weight decay        & 0.05    & 0.05      \\
Rand Augment        & 9/0.5   & 9/0.5       \\
Mixup               & None     & 0.8   \\
Cutmix              & None    & 1.0      \\
Label smoothing     & 0.1     & None     \\ \hline
\end{tabular}
\end{table}

\begin{table}[h]
\caption{Configurations of different QSD-Transformer models, which is similar to our strong baseline Spike-driven Transformer v2 \citep{yao2024spike}.}
\label{table_imagenet_config}
%{5pt}
% \tabcolsep=0.08cm
\begin{tabular}{c|c|ccc|ccc}
\hline
stage & \# Tokens & \multicolumn{3}{c|}{Layer Specification}& \multicolumn{1}{c|}{1.8M} & \multicolumn{1}{c|}{3.8M} &6.8M \\ \hline

\multirow{12}{*}{1} & \multirow{6}{*}{$\dfrac{H}{2}$ \texttimes $\dfrac{W}{2}$}   & \multicolumn{2}{c|}{\multirow{2}{*}{Downsampling}} & Conv & \multicolumn{3}{c}{7x7 stride 2}
\\ \cline{5-8} 
& & \multicolumn{2}{c|}{} & Dim & \multicolumn{1}{c|}{32}  & \multicolumn{1}{c|}{48}  & 64  
\\ \cline{3-8} 
& & \multicolumn{1}{c|}{\multirow{4}{*}{\begin{tabular}[c]{@{}c@{}}Conv-based\\      SNN block\end{tabular}}}        & \multicolumn{1}{c|}{\multirow{2}{*}{SepConv}}      & DWConv     & \multicolumn{3}{c}{7x7 stride 1}                          
\\ \cline{5-8} 
& & \multicolumn{1}{c|}{} & \multicolumn{1}{c|}{} & FC ratio  & \multicolumn{3}{c}{2} 
\\ \cline{4-8} 
& & \multicolumn{1}{c|}{} & \multicolumn{1}{c|}{\multirow{2}{*}{Channel Conv}} & Conv       & \multicolumn{3}{c}{3x3 stride 1}                        
\\ \cline{5-8} 
& & \multicolumn{1}{c|}{} & \multicolumn{1}{c|}{} & Conv ratio & \multicolumn{3}{c}{4}                \\ \cline{2-8} 
& \multirow{6}{*}{$\dfrac{H}{4}$ \texttimes $\dfrac{W}{4}$}   & \multicolumn{2}{c|}{\multirow{2}{*}{Downsampling}} & Conv       & \multicolumn{3}{c}{3x3 stride 2}            \\ \cline{5-8} 
& & \multicolumn{2}{c|}{} & Dim        & \multicolumn{1}{c|}{64}  & \multicolumn{1}{c|}{96}  & 128 
\\ \cline{3-8} 
& & \multicolumn{1}{c|}{\multirow{4}{*}{\begin{tabular}[c]{@{}c@{}}Conv-based\\      SNN block\end{tabular}}}        & \multicolumn{1}{c|}{\multirow{2}{*}{SepConv}}      & DWConv     & \multicolumn{3}{c}{7x7 stride 1}                          
\\ \cline{5-8} 
& & \multicolumn{1}{c|}{} & \multicolumn{1}{c|}{}                              & FC ratio  & \multicolumn{3}{c}{2}                                     
\\ \cline{4-8} 
& & \multicolumn{1}{c|}{} & \multicolumn{1}{c|}{\multirow{2}{*}{Channel Conv}} & Conv       & \multicolumn{3}{c}{3x3 stride 1}                        \\ \cline{5-8} 
& & \multicolumn{1}{c|}{} & \multicolumn{1}{c|}{} & Conv ratio & \multicolumn{3}{c}{4}                                     
\\ \hline
\multirow{7}{*}{2}  & \multirow{7}{*}{$\dfrac{H}{8}$ \texttimes $\dfrac{W}{8}$}   & \multicolumn{2}{c|}{\multirow{2}{*}{Downsampling}} & Conv       & \multicolumn{3}{c}{3x3 stride 2}                          
\\ \cline{5-8} 
& & \multicolumn{2}{c|}{} & Dim & \multicolumn{1}{c|}{128} & \multicolumn{1}{c|}{192} & 256 
\\ \cline{3-8} 
& & \multicolumn{1}{c|}{\multirow{5}{*}{\begin{tabular}[c]{@{}c@{}}Conv-based\\      SNN block\end{tabular}}}        & \multicolumn{1}{c|}{\multirow{2}{*}{SepConv}}      & DWConv     & \multicolumn{3}{c}{7x7 stride 1}                          
\\ \cline{5-8} 
& & \multicolumn{1}{c|}{} & \multicolumn{1}{c|}{}                              & FC ratio  & \multicolumn{3}{c}{2}                                     
\\ \cline{4-8} & & \multicolumn{1}{c|}{} & \multicolumn{1}{c|}{\multirow{2}{*}{Channel Conv}} & Conv       & \multicolumn{3}{c}{3x3 stride 1}          \\ \cline{5-8} 
& & \multicolumn{1}{c|}{} & \multicolumn{1}{c|}{}                              & Conv ratio & \multicolumn{3}{c}{4}                                     
\\ \cline{4-8} 
& & \multicolumn{1}{c|}{} & \multicolumn{2}{c|}{\# Blocks} & \multicolumn{3}{c}{2}                                    
\\ \hline
\multirow{5}{*}{3}  & \multirow{5}{*}{$\dfrac{H}{16}$ \texttimes $\dfrac{W}{16}$} & \multicolumn{2}{c|}{\multirow{2}{*}{Downsampling}} & Conv       & \multicolumn{3}{c}{3x3 stride 2}                         
\\ \cline{5-8} 
& & \multicolumn{2}{c|}{} & Dim        & \multicolumn{1}{c|}{256} & \multicolumn{1}{c|}{384} & 512 
\\ \cline{3-8} 
& & \multicolumn{1}{c|}{\multirow{3}{*}{\begin{tabular}[c]{@{}c@{}}Transformer-based\\      SNN block\end{tabular}}} & \multicolumn{1}{c|}{SDSA}                          & RepConv    & \multicolumn{3}{c}{3x3 stride 1}                          
\\ \cline{4-8} 
& & \multicolumn{1}{c|}{} & \multicolumn{1}{c|}{Channel FC}                   & FC ratio  & \multicolumn{3}{c}{4}                                     
\\ \cline{4-8}  
& & \multicolumn{1}{c|}{} & \multicolumn{2}{c|}{\# Blocks} & \multicolumn{3}{c}{6}                                     
\\ \hline
\multirow{5}{*}{4}  & \multirow{5}{*}{$\dfrac{H}{16}$ \texttimes $\dfrac{W}{16}$} & \multicolumn{2}{c|}{\multirow{2}{*}{Downsampling}} & Conv       & \multicolumn{3}{c}{3x3 stride 1}                         
\\ \cline{5-8} 
& & \multicolumn{2}{c|}{}  & Dim & \multicolumn{1}{c|}{360} & \multicolumn{1}{c|}{480} & 640 
\\ \cline{3-8} 
&  & \multicolumn{1}{c|}{\multirow{3}{*}{\begin{tabular}[c]{@{}c@{}}Transformer-based\\      SNN block\end{tabular}}} & \multicolumn{1}{c|}{SDSA}                          & RepConv    & \multicolumn{3}{c}{3x3 stride 1}                          
\\ \cline{4-8} 
& & \multicolumn{1}{c|}{} & \multicolumn{1}{c|}{Channel FC}                   & FC ratio  & \multicolumn{3}{c}{4}                                     
\\ \cline{4-8} 
& & \multicolumn{1}{c|}{} & \multicolumn{2}{c|}{\# Blocks} & \multicolumn{3}{c}{2}                                    
\\ \hline
\end{tabular}
\end{table}

\subsection{COCO experiments}\label{appendix_training_COCO}
%在COCO实验中，我们在imagenet-1k上预训练QSD-Transformer作为backbone，然后加入MaskR-CNN后在COCO数据集上微调24个epochs来得到最后的模型。在微调阶段，我们对训练数据和测试数据resize并裁剪为1333*800，此外，我们还对训练数据做了随机的水平翻转来提高泛化性。我们使用AdamW优化器并初始化学习率为1e-4，学习率使用0.9的多项式衰减。
%为了减小模型大小，我们使用更轻量化的Yolov5作为分类器进行了进一步实验。我们对Yolov5做了改进，使其更符合SNN的特性，实验结果如表\ref{}所示。与现阶段的SOTA结果相比，我们的方法能实现更高的精度。
The COCO dataset aims at scene understanding, primarily extracted from complex everyday scenes, where objects in images are precisely localized through accurate segmentation. The COCO dataset comprises 118K training images and 5K validation images. In the COCO experiments, we pre-trained the QSD-Transformer on ImageNet-1k as the backbone, and then fine-tuned it on the COCO dataset for 24 epochs with the Mask R-CNN as detector to obtain the final model. During the fine-tuning stage, we resized and cropped the training and test data to 1333x800. Additionally, we applied random horizontal flipping and resize with a ratio of 0.5 to the training data. The batch size was set to 12. We used the AdamW optimizer with an initial learning rate of 1e-4, and the learning rate was decayed polynomially with a power of 0.9. We conducted training on four 40GB A100 GPUs for a duration of 26 hours. 
% The results of our method on the COCO dataset are shown in Fig. \ref{fig:coco}.

\subsection{ADE20K experiments}\label{appendix_training_ADE20K}
%我们使用在imagenet上预训练的QSD-Transformer作为backbone结合FPN来进行segmentation实验。我们使用Xavier初始化新添加的参数，并在ADE20K数据集上以20的batchsize训练160K个iterations。我们使用AdamW优化器并初始化学习率为1e-4，学习率使用0.9的多项式衰减。
The ADE20K semantic segmentation dataset comprises over 20K training and 2K validation scene-centric images meticulously annotated with pixel-level object and object parts labels, fostering a comprehensive understanding of complex scenes. It encompasses a total of 150 semantic categories, encompassing elements such as sky, road, and grass, as well as discrete entities like person, car, and bed. We also used the QSD-Transformer pre-trained on ImageNet-1K as the backbone combined with FPN for segmentation experiments. The newly added parameters were initialized using Xavier initialization, and the model was trained on the ADE20K dataset with a batch size of 20 for 160K iterations. We utilized the AdamW optimizer with an initial learning rate of $1 \times 10^{-4}$, and the learning rate was decayed polynomially with a power of 0.9. During the initial 1500 iterations, we employed linear decay to warm up the model. The training process was executed on four 40GB A100 GPUs and lasted for 25 hours.
% The results of our method on the ADE20K dataset are shown in Fig. \ref{fig:segmentation}.

\subsection{Transfer learning}\label{appendix_transfer_learning}
%在迁移学习实验中，我们首先加载在imagenet上预训练好的参数，然后根据数据集类别数量替换最后一层全连接层，例如在cifar100实验中将1000-FC替换为100-FC。在微调训练中，我们首先对训练数据进行数据增强，包括mixupcutmix以及label smoothing。我们设置batchsize为128，并使用Adamw优化器with a weight decay of 0.01. 我们在100个epochs中使用初始化为1e-4的余弦衰减学习率。
%此外，对于神经形态数据集，我们还使用了额外的数据预处理。首先我们将事件流分为$T$个切片，其中T为设置的时间步长，每个切片有相同数量的事件。然后我们分别在每个切片中将事件压缩为三通道的帧，三通道分别为positive、negative、all。这样，我们就将事件流转换为T个时间步的数据。我们还对处理后的事件数据使用了数据增强方法，similar to \citep{li2022neuromorphic}
We performed transfer learning experiments on the static image classification datasets CIFAR10/100 and the neuromorphic classification dataset CIFAR10-DVS. The CIFAR10/100 datasets each have 50,000 training and 10,000 test images with a resolution of \(32\times32\). CIFAR10-DVS consists of 10K event streams created by capturing CIFAR10 images using a DVS camera.

In these experiments, we first loaded pre-trained ImageNet-1K checkpoints and replaced the final fully connected layer to match the number of classes in each dataset (e.g., replacing the 1000-FC with 100-FC for CIFAR-100). During fine-tuning, we applied data augmentations like mixup, cutmix, and label smoothing. We used a batch size of 128, the AdamW optimizer with a weight decay of 0.01, and a cosine-decay learning rate schedule starting at \(1 \times 10^{-4}\) over 100 epochs. The experiments were run on a single 32GB V100 GPU, taking 12 hours for CIFAR-10 and CIFAR-100, and 10 hours for CIFAR-10-DVS.

For CIFAR10-DVS, we added preprocessing steps: dividing the event stream into \(T\) slices, each with an equal number of events, and compressing these into three-channel frames representing positive, negative, and all events, transforming the event stream into \(T\) frames. We also applied data augmentation to the processed event data, as described in \citep{wang2023masked,shi2024spikingresformer}.
\begin{figure}[htbp]
    \centering
    \includegraphics[width=1\textwidth]{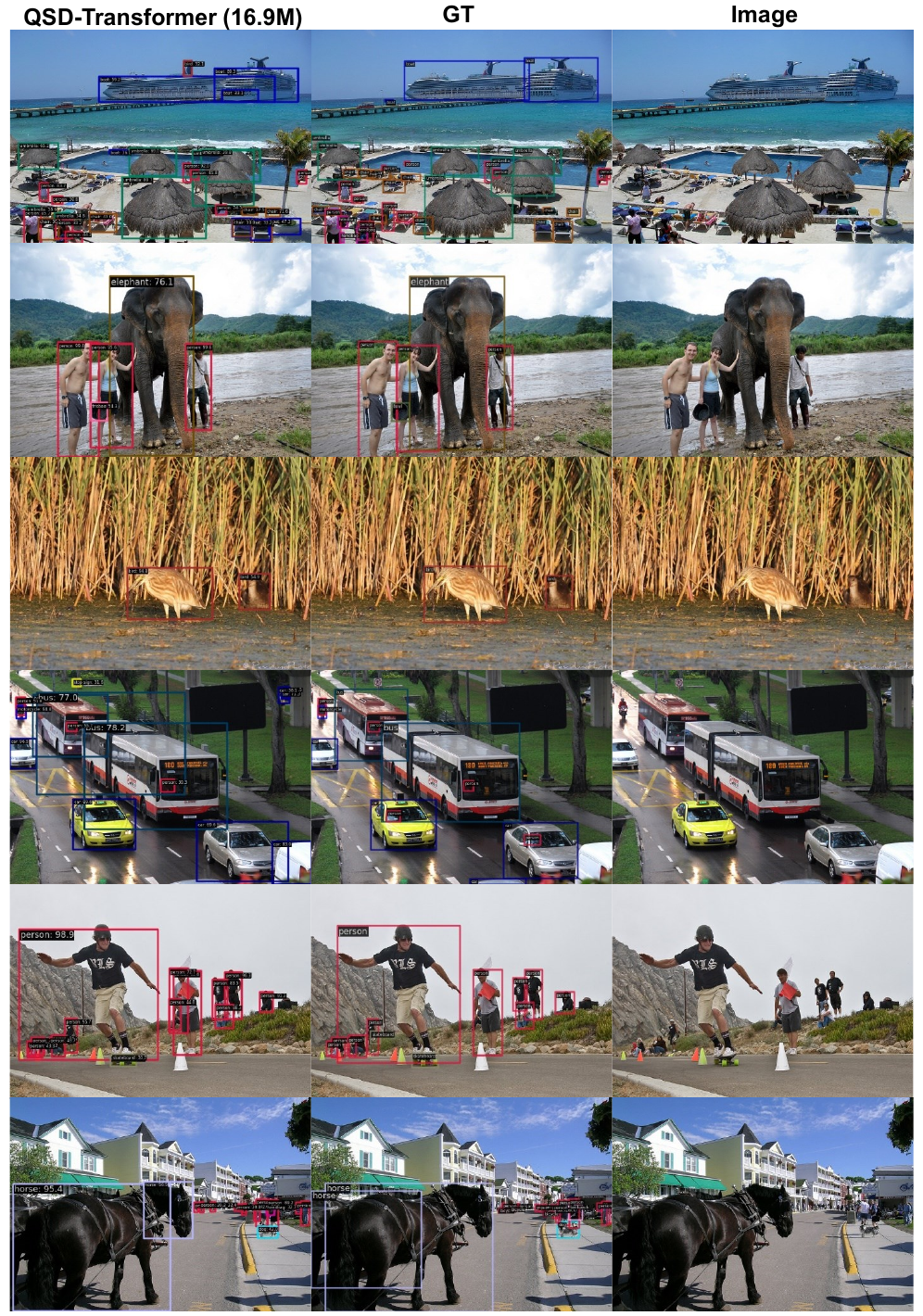}
    \caption{Visualization of results on COCO dataset. Our QSD-Transformer excels in the target detection task.}
    \label{fig:coco}
\end{figure}

\begin{figure}[htbp]
    \centering
    \includegraphics[width=1\textwidth]{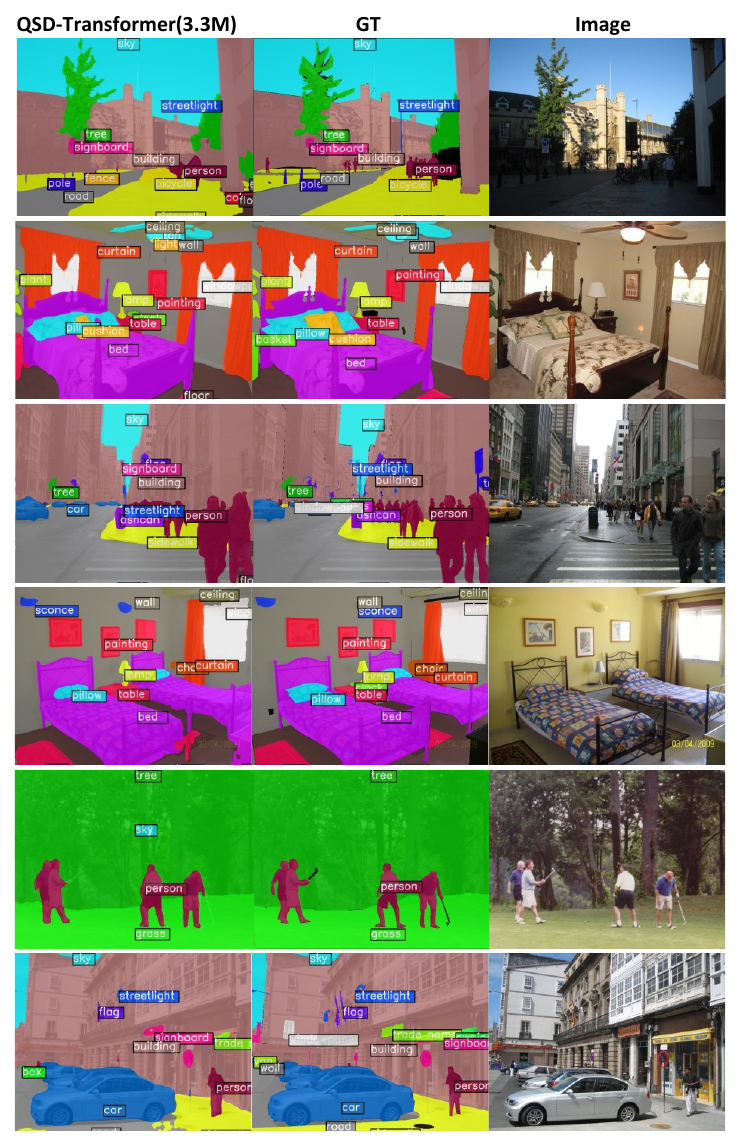}
    \caption{Visualization of results on ADE20K dataset. Our QSD-Transformer excels in the segmentation task.}
    \label{fig:segmentation}
\end{figure}

\end{document}